\documentclass[a4paper,fleqn]{cas-dc} 

\usepackage[authoryear]{natbib}
\usepackage{mathptmx}
\usepackage{multirow}
\usepackage{caption}
\usepackage{threeparttable}
\usepackage{balance}

\usepackage{amsmath,amssymb,amsfonts}
\usepackage{algorithmic}
\usepackage{graphicx}
\usepackage{textcomp}
\usepackage{balance}
\usepackage{mathtools}  

\ExplSyntaxOn
\cs_gset:Npn \__first_footerline:
{
  \group_begin:
  \small
  \sffamily
  \ifnum\theblind>0\relax
  \else
    \__short_authors:
  \fi
  \group_end:
}
\ExplSyntaxOff

\begin{document}

\let\WriteBookmarks\relax
\def\floatpagepagefraction{1}
\def\textpagefraction{.001}

\shorttitle{Evaluating LLMs for Implicit Sentiment Analysis of Product Desirability}

\shortauthors{S. Weitl-Harms and J. Hastings}

\title [mode = title]{Evaluating LLM Usage for Efficient and Explainable Numerical and Classified Implicit Sentiment Analysis of Product Desirability}

\author[1]{Sherri Weitl-Harms}[type=editor,
                        auid=000,bioid=1,
                        role=Researcher,
                        orcid=0000-0002-3653-2928]

\cormark[1]
\ead{sherriweitlharms@creighton.edu}

\credit{Conceptualization, Formal analysis, Funding acquisition, Investigation, Methodology, Project administration, Resources, Validation, Visualization, Writing – original draft, Writing – review \& editing}

\affiliation[1]{organization={Creighton University},
    addressline={2500 California Plz}, 
    city={Omaha, NE},
    postcode={68178}, 
    country={United States}}

\author[2]{John Hastings}[type=editor,
                        auid=000,bioid=1,
                        role=Researcher,
                        orcid=0000-0003-0871-3622]

\credit{Conceptualization, Data curation, Investigation, Methodology, Software, Writing – original draft, Writing – review \& editing}

\affiliation[2]{organization={The Beacom College of Computer \& Cyber Sciences, Dakota State University},
    addressline={820 N Washington Ave.}, 
    city={Madison, SD},
    postcode={57042},
    country={United States}}

\cortext[cor1]{Principal corresponding author}

\date{June 6, 2026}
\begin{abstract}
Qualitative product feedback can reveal nuanced user experiences, but its implicit sentiment is difficult to measure. This paper presents a scalable and interpretable framework that uses large language models (LLMs) to quantify product desirability from such data. Using two Product Desirability Toolkit (PDT) datasets from ZORQ and CARMA comprising 106 respondent term groupings with gold-standard human annotation, zero-shot continuous numerical sentiment scoring and categorical sentiment classification are evaluated without relying on explicit review scores. Across the datasets, LLMs generated numerical sentiment scores directly from qualitative responses and closely matched expert labels, achieving Pearson correlations up to 0.97 and classification accuracy up to 94\%. LLMs maintained robustness even when handling data presented in multiple forms and consistently expressed high confidence. In contrast, lexicon-based and transformer baselines did not produce statistically significant results. Among the models tested, GPT-4o-mini achieved performance comparable to larger models at 94\% lower cost, supporting scalable deployment. 

The framework also incorporates model confidence ratings and human-readable rationale explanations (xAI),  improving interpretability, transparency, and trust while supporting practical use in product satisfaction assessment. In general, using the PDT tool as a survey method along with a cost efficient LLM for sentiment analysis has the potential to provide for product evaluation with results that are rich in terms of sentiment scores (both numerical and classified sentiment) and in terms of the high-level user impressions of the product that can be used to identify ideas for product development and improvement, as well as marketing ideas for target audiences. 
\end{abstract}


\maketitle

\section{Introduction}\label{introduction}
In product development, understanding implicit user sentiment is crucial not only for creating products that truly appeal to their intended audience, but also for improving and marketing those products effectively. Sentiment analysis, a field dedicated to automatically extracting emotional reactions expressed by users \citep{Zhang2018}, holds the potential to provide deeper insights into user perceptions, satisfaction, and overall product desirability. Sentiment information can help information retrieval systems intelligently perform a more accurate and responsive retrieval and recommendation of relevant product information for users \citep{Buonaiuto}.

Recent advancements in natural language processing (NLP), particularly the development of large language models (LLMs), have opened new possibilities for sentiment analysis, particularly for qualitative data. These models have demonstrated capabilities in various tasks, including zero-shot sentiment classification \citep{Ren11413928,Fazzi, Xiao2024,ZHANG2026104703}. The increase in usage raises significant questions about the transparency and interpretability of the predictions made by these LLMs 
\citep{Tsai}. Oftentimes, these models focus on single aspects of affective classification tasks (e.g., sentiment polarity or categorical emotions), and overlook quantitative elements (e.g., sentiment strength or emotion intensity) \citep{Liu10,Zeng,Jia10.1145/3799234}.

Finding an efficient (in terms of price and power consumption) alternative to larger commercial LLMs is an important consideration, especially given the 
environmental impacts of training and running AI models 
\citep{strubell2019energy,schwartz2020green,kaack2022aligning}. \citet{crawford2024generative} argues that immediate action is needed to limit the environmental impacts of AI. Among commercial LLMs, GPT4o-mini is designed to be a smaller, more cost-efficient model in comparison to larger models \citep{gpt4omini}, which should normally translate to a reduced carbon footprint. However, OpenAI has not commented specifically on this aspect or the nature of the resources (and thus the associated environmental impacts) consumed in producing GPT4o-mini relative to other models.

Often, user reviews or expressions on social media are used  as a basis for sentiment analysis.  Understanding implicit sentiment, a common linguistic phenomenon where accurate judgment often requires common sense or domain knowledge, is 
difficult \citep{Liu11411769,Liu10}. Quantifying implicit user sentiment remains a significant challenge, particularly when explicit user ratings or reviews are unavailable \citep{zhou2021implicit}. Volo (\citeyear{Volo}) highlighted the need for a user-centered design (UCD) method
capable of unobtrusively accessing individuals’ experience
while minimizing investigator and selection biases in its
translations, measurements and analyses. These requirements
demand an approach that is structured and controlled yet also
flexible and open-ended \citep{packer}. A major barrier to UCD is the
relative lack of formal mechanisms to translate individual user
“voices” into the design of distinct product attributes that
reflect divergent preferences \citep{Salvador}. Devising intelligent systems
that can identify users' unique needs at scale and translate
them into attribute-level design feedback and recommendations
is essential for effective UCD processes \citep{HAN2021114604}.

To address situations in which sentiment data is lacking, tools such as surveys and the Microsoft Product Desirability Toolkit (PDT) can be used to evaluate user experiences. The PDT~\citep{Benedek,benedek2002a} is recognized as a valuable qualitative tool for evaluating user experience and satisfaction. However, while it excels at gathering rich qualitative data, it lacks inherent quantitative abilities \citep{Sauro}. 

The sections that follow provide the research objectives, provide background information on the PDT and sentiment analysis techniques, detail the methodology, present results, discuss their implications, and outline directions for future work.

\subsection{Research Objectives}
This research provides a zero-shot framework for numerical implicit sentiment extraction from qualitative user experience data. The approach is to input PDT qualitative data into LLMs to provide numeric sentiment scores as well as confidence explanations, for evaluation and  application to decision making. This research aims to bridge the gap between qualitative sentiment data and quantitative analysis by applying recent LLMs and other sentiment analysis techniques to PDT data. 

The overall goal is to derive meaningful quantitative and categorical sentiment insights from qualitative user responses. In addition, by comparing LLMs with existing approaches, the research seeks to identify the most effective methods for quantifying and classifying implicit user sentiment in software desirability evaluations. The following research questions guide the study:

\begin{enumerate}
   \item[1)] How do various LLMs compare to known tools, VADER and Twitter-RoBERTa-Base-Sentiment, 
   in accuracy for determining implicit user sentiment expressed in qualitative PDT data?
     \item[2)] How accurate are the LLMs for numerical sentiment analysis?
     \item[3)] What are the LLM costs to conduct the numerical sentiment analysis?
      \item[4)] How accurate are the LLMs for categorical sentiment analysis? 
    \item[5)] How much data do LLMs such as GPT4o and GPT4o-mini need to still maintain a high level of accuracy? 
    \item[6)] Does having the LLMs provide confidence ratings and explanations add value to quantifying implicit user sentiment?
\end{enumerate}

\subsection{Research Contributions}
This research develops a zero-shot generalizable pipeline for converting qualitative user experience data into quantitative sentiment, using Large Language Models (LLMs), without the need for explicit review scores. The novel framework of using the PDT tool as the survey method along with LLM sentiment analysis provides for product evaluation where higher-level impressions of that product are desired. The PDT survey method is quick and easy to construct and distribute and can elicit candid non-biased feedback from participants about a product. The combined use of the PDT and LLM sentiment analysis also minimizes investigator and selection biases, while providing a controlled and structured approach that is also open-ended and flexible. 

To verify that this framework works well, the research provides a comparison with traditional categorical sentiment classifications that demonstrates the LLMs achieve strong agreement with expert labels, and includes model confidence and rationale explanations for transparency and enhanced interpretability (xAI). Finally, the research addresses the cost of the models included in the framework, by conducting an analysis of cost-performance tradeoffs. 

Results show that LLMs matched expert-labeled numerical and categorical sentiment with high accuracy (Pearson correlations up to 0.97, classification accuracy up to 94\%) and far outperformed legacy tools, which failed to produce statistically significant results. GPT4o-mini achieved performance comparable to the other LLMs, at a 94\% lower cost. LLMs maintained robustness even when handling data presented in multiple forms and consistently expressed high confidence with human-readable justifications. These qualities support interpretability and trust, enabling use in many product satisfaction contexts. Overall, this work advances a scalable and interpretable framework for capturing implicit quantitatively scored product desirability.

\section{Related Works}\label{background}

Understanding users' implicit sentiment related to product usability is important for many reasons, including improving and marketing products, but it is difficult to measure \citep{Zhang2018, hastings2024}. Often, user reviews or expressions on social media are used  as a basis for sentiment analysis.  Concepts like `enjoyment', `fun' or a product's desirability for purchase or use are not effectively captured by traditional usability studies~\citep{Benedek}. 

Measuring desirability along with usability offers a layer of qualitative impact, hence providing a much better picture of the user experience \citep{Carrabina}. "Usable" refers to the ease of use and efficiency of a product, while "desirable" refers to the emotional and psychological appeal of a product, often related to its aesthetic, branding, and perceived value. According to ISO 9241-11 \citep{iso924111}, usability is  ``the extent to which a product can be used by a user to achieve a goal with effectiveness, efficiency, and satisfaction.''  
Measuring usability is complex because it is intrinsic to the system or object under evaluation \citep{Carrabina}. 
In product development, understanding implicit user sentiment of product desirability is crucial for creating products that truly appeal to their intended audience, especially in user-centered design (UCD) \citep{Marina}. 

\subsection{Product Desirability Toolkit (PDT)}
To address situations in which sentiment data or user review data is lacking, tools such as surveys and the Microsoft Product Desirability Toolkit (PDT)~\citep{Benedek,benedek2002a} can be used to evaluate user experiences. 
The PDT is a qualitative analysis tool used to evaluate user experience and satisfaction with products, such as software \citep{Barnum2010, Barnum, Booth2013EndUserEO, Hastings, Li2014, Tullis, Veral, Weitl}. It aims to ``understand the illusive, intangible aspect of desirability resulting from a user’s experience with a product'' \citep{Barnum}. The PDT asks users to select five adjectives from a given set that best describe their feelings about the experience, and provide an optional explanation for each word choice. By gathering this group of word/explanation pairs, called a PDT Respondent Term Grouping (PRTG), the approach is designed to capture rich, qualitative data about user experiences and perceptions.

The advantages of using the PDT are ``1) it aims to avoid a bias toward the positive found in typical questionnaires (e.g., it has been found that if a respondent thinks that a survey intends to assess the quality of a product, they are likely to provide more positive answers about quality) and 2) it is able to more effectively uncover constructive negative criticisms in the guided interview'' \citep{Hastings}.

The PDT is described as the closest tool that uses ``psychometric theory to create a user experience (UX)-relevant measure of product or service desirability'' \citep{Sauro}. The design of the PDT prompts users to tell a revealing story of their experience as they comment on their word choice \citep{Barnum2010}, thereby providing a rich set of qualitative data related to the implicit product desirability. 

While the PDT is a great tool for capturing qualitative product satisfaction, it is a poor quantitative tool by itself \citep{Sauro}. The PDT provides a way to triangulate findings from other feedback mechanisms, with potential to produce more meaningful and substantive results of user experiences \citep{Barnum2010}, but it is necessary to think of improvements on the original method \citep{Veral}. The PDT text from respondent word/comment groupings is ripe for sentiment analysis.

\subsection{Sentiment Analysis}

Sentiment analysis has emerged as 
a field dedicated to automatically extracting emotional reactions expressed by users \citep{Zhang2018,Liu11411769,LIU2026104746,Jia10.1145/3799234}.
It covers diverse applications, such as guiding policy-making, optimizing product development \citep{Bharadwaj}, maintaining brand image, and managing crisis communication \citep{kaur,Tsai, Sharma}.

For example, businesses gain 
value from sentiment analysis for many purposes, including understanding user perception, market research, financial analysis \citep{Xiao2024,Deng,Abdelsamie} and  customer satisfaction \citep{Tsai,Venkata,Chinnalagu,Qiao}. In the communications field, sentiment analysis with few-shot learning was used to detect hate speech by first employing publicly available sentiment datasets to train a sentiment analysis model, and then fine-tuning the model by merging sentiment prompts with hate speech prompts \citep{Wang10.1145}.  Sentiment analysis was used to evaluate impact on individual's communication, information access, and daily activities  during an internet outage in Bangladesh \citep{Tahmidul}. 
Sentiment analysis has even been used to study LLMs' ability to  assess their affective expressive range in terms of arousal and valence during extended dialogues \citep{Fazzi}.

Sentiment analysis employs machine learning algorithms, statistical methods, and NLP to identify patterns and trends among opinions, emotions, and attitudes expressed in text, and subsequently classifies them into categories or sentiment scores \citep{Tsai,Medhat,Klongdee}. Quantifying implicit user sentiment is challenging, but important for understanding user experiences \citep{Liu10}.

Traditional sentiment analysis techniques often rely on lexicon-based approaches or machine learning models trained on labeled data. However, these methods can struggle with context-dependent sentiments and implicit expressions of user feelings \citep{dang2020sentiment}, which are common in PDT responses.

There are regression-based approaches to sentiment analysis \citep{liang,CHIHAB2022}. Logistic regression is commonly used for sentiment analysis due to its simplicity, interpretability, and proven effectiveness in classification, but is not as useful for quantifying sentiment. Linear regression has been used for quantifying sentiment, but regression relies on explicit seed words or user ratings, limiting its use for implicit sentiment analysis.  For example, linear regression has been used to generate quantified sentiment scores for words, and then determine the intensity of a given word from a set of similar words \citep{liang}. Quantifying phrase sentiment was then done by aggregating the numeric intensity for the words in the phrases. This approach can require seed words, and suffers from many of the issues described below, especially multipolarity. Linear regression models have also been used for sentiment classification \citep{CHIHAB2022}.

Another sentiment analysis approach is embedding-based sentiment analysis, which represents words, sentences, or documents as dense, numerical vectors (embeddings) in a high-dimensional space, with similar emotional meanings positioned close together. By analyzing the positions of these vectors, systems classify the sentiment \citep{hamdhana}. However, this approach requires explicit ratings to complete the sentiment analysis and provides classification rather than quantitative sentiment.

LLMs have shown promising results in various NLP tasks, including their ability to gauge sentiment effectively \citep{Tsai, Susnjak, Krugmann, Akihito, Kazi, hastings2024, gelman2025scalable}. LLMs, such as GPT models, are trained on vast amounts of text data and can generate human-like text based on input prompts. Similar to sentiment analysis, LLMs have also been used to recognize an author’s stance in written text, including zero-shot stance detection (ZSSD) that recognizes the stance towards unknown targets \citep{UPADHYAYA2025104223}. They are also used in interactive recommender systems for providing personalized content to users \citep{LIU2026104746}. Starting in late 2023, GPT (gpt-3.5-turbo-0301) demonstrated impressive zero-shot capabilities in sentiment classification tasks, and could serve as a universal and well-behaved sentiment analyzer \citep{wang2023chatgpt}.  
Generative AI has pioneered zero-shot content analysis, such as automated textual analysis \citep{Krugmann}.
However, early versions of LLMs generally performed poorly on implicit sentiment analysis and domain-specific training was needed to improve performance \citep{wang2023chatgpt}. The application of newer LLMs to implicit sentiment analysis offers new possibilities for understanding user experiences and product desirability \citep{Ren11413928, Lai}. The ability of LLMs to understand context and capture nuanced and implicit sentiments could potentially overcome some of the limitations of traditional sentiment analysis techniques for quantifying implicit sentiment when applied to PDT data.

The most common approach to sentiment analysis using LLMs is to classify sentiment into three categories (positive, neutral and negative) or five categories (strongly positive, positive, neutral, negative, and strongly negative) \citep{Liu10, Deng, Zeng,Abdelsamie}. A confusion matrix, accuracy, precision, recall and F1-scores are calculated for comparison \citep{Chinnalagu}.  Most approaches also use data sets that include test data with known values (such as five-star ratings) for analysis.  A comparison of accuracy when classifying sentiment into two to seven different categories was done in \citep{Zeng}, which found that accuracy and F1-scores decreased as the number of categories increased.

\subsection{Implicit Sentiment Analysis (ISA)}

Identifying user sentiment in product reviews when explicit user ratings or reviews are unavailable, requires an evaluation of the implicit sentiment \citep{Ren11413928}. Mapping user needs onto design specifications through sentiment analysis with deep learning tools \citep{WANG2018145} has promise. However, document-level sentiment classification is unable to capture attribute-level information \citep{HAN2021114604}. Additionally, these methods require manually labeled data for training, and often classify sentiment into predefined categories (such as positive, neutral, and negative sentiment) which have limited implications for UCD designers \citep{HAN2021114604}. Providing quantified implicit sentiment is much more challenging, but extremely useful. In general, the supervised nature of most sentiment classification approaches limits their practicality as they require extensive manual data labeling and annotation for training \citep{HAN2021114604}.

Due to the relatively subtle emotional color of implicit sentiment text and its relatively unclear representation in the text, traditional sentiment analysis methods cannot achieve good results for ISA \citep{Hu11100157}. Most traditional classifiers find it challenging to address ISA due to limited
contextual information and insufficient reasoning skills, and address ISA superficially \citep{Lai}. In general, when early LLMs were directly deployed for ISA without training, their reasoning capacity was not fully harnessed, resulting in suboptimal results \citep{Lai}. 

ISA research shifted toward causal reasoning and prompt-based approaches, motivated by the success of LLMs \citep{Liu11411769}. For categorical implicit sentiment analysis, emotion-aware generation \citep{Ouyang} and prompt-based methods such as THOR \citep{Fei}, RVISA \citep{Lai} and CAPITAL \citep{Ren11413928} use enhanced chain of thought reasoning to infer implicit emotions by analyzing nuanced language. Aspect-based sentiment analysis links sentiments to specific aspects within a text  using manual labeling of implicit terms and aspects \citep{Rahim11138369,Liu11411769}.  Despite advances, current ISA methods remain constrained by their inability to perform multistep reasoning and their dependence on manually engineered features or annotated data sets \citep{Liu11411769}.  \citet{Liu11411769} addressed these issues by using progressive prompting strategies.

Numerical implicit sentiment analysis of PDT data using various LLMs was introduced in \cite{weitl2024analyzing}. Their results show a promising novel method to quantitatively measuring implicit user desirability with statistically significant results, and provide a baseline for further numerical implicit sentiment analysis. Their combined use of the PDT and LLM sentiment analysis limits investigator and selection biases as suggested by \cite{Volo}, while the method is controlled and structured but also open-ended
and flexible as suggested by \cite{packer}.  
However, there is need for both a comparison of conducting zero-shot numerical and categorical sentiment analysis on product desirability data and a comparison of the cost-effectiveness of using LLMs for sentiment analysis.

\subsubsection{Issues with Sentiment Analysis}

Sentiment analysis has known limitations, which include:
\begin{itemize}
    \item Sarcasm Detection - not likely to be an issue with PDT reviews - as the PDT word bank keeps the user expressions focused on the context of the product being reviewed.
    \item Negation Detection - this is also not likely to be an issue as the PDT word bank keeps the user expressions focused on the context of the product being reviewed, and asks the user to express why a certain word was selected.
\item Word Ambiguity - where the polarity of words depends on the context. The PDT word bank keeps the user expressions focused on the context of the product being reviewed, and is generally known as unambiguous words~\citep{Veral}, and was selected for the purpose of allowing the user to express related sentiment~\citep{benedek2002a}.
\item Multipolarity - where a user selects both positive and negative sentiment words, resulting in a neutral sentiment score of the overall review. This will be addressed in future research, when looking at the dimensional/aspect-level of sentiment analysis \citep{Tanja1,WangXu,Maag,ZhuL,Jia10.1145/3799234}.
\end{itemize}

\section{Methods}
This research focuses on sentiment analysis of PDT survey datasets, each containing five words and explanations from respondents. The goal is to evaluate the effectiveness of various sentiment analysis technologies, particularly LLMs, on this data. 
This research performs scaled (ranging from 0-1) numerical sentiment analysis, as well as classifies sentiment into the traditional three classes (positive, neutral, and negative) as done in \citep{Krugmann}, and five classes (very positive, positive, neutral, negative, and very negative). Numerical analysis is more challenging, but offers significant advantages, especially in that it provides more detail and insights into the magnitude of sentiment, which can help decision makers make better decisions regarding product desirability. 

\subsection{Data Collection}

The data sets for this study were previously collected \citep{ Weitl} from users of CARMA \citep{Hastings}, a grasshopper infestation rangeland management system and ZORQ (Zero Operator ReQuired) \citep{Hastings2022}, a gamification framework utilized in undergraduate computer science education. 

CARMA, short for CAse-based Rangeland grasshopper Management Advisor has been successfully used since 1996 \citep{Hastings2002}. In order to gain non-biased feedback about the quality of CARMA’s interface, a group of novice users not previously familiar with CARMA were surveyed. The participants were part of the general student population at the regional university. The study received ethical approval from the university, and students were given informed consent to opt-in to completing the survey. The survey was distributed en masse via email to these students in May 2009. The survey received 89 responses. Initial analysis of the comments revealed that 37 respondents (41\% of the total responses) did not provide valid responses. For example, a certain percentage of participants were not serious about the survey and filled in nonsense answers in order to enter the drawing for the pizza (e.g., comments such as “zscfsd sdds...”). Another group of participants were confused about the survey and rather than provide responses about CARMA’s interface through term selection and comments, chose to provide commentary on the domain (i.e., grasshopper management) or the applicability of the domain to themselves, e.g., some participants selected the term irrelevant because “I don’t see what i have to do with grasshoppers” or “I don’t live on a farm/ranch and I don’t have a grasshopper infestation.”. After filtering for invalid survey responses, there were 56 valid PRTG responses (with five PDT terms each) \citep{Hastings}. The participants were not given advance warning as to how a selected term would be used or that a follow-up comment would be required in an attempt to reduce the effect of bias on the selection of any term. Because this was an anonymously online survey, to implement the second phase of the PDT method, each participant was asked to comment on each of the five terms selected in the online survey itself, rather than conducting a face-to-face follow-up interview session \citep{Hastings}.

ZORQ is a game development framework system in which spaceships navigate a 2D game universe filled with objects which affect a ship either positively or negatively. It is
used by students in data structures and algorithms courses.  An anonymous online survey was distributed during the fall 2021, fall 2020, fall 2019, fall 2018, and fall 2017 semesters as described by \citep{Hastings2022} at a regional university that used ZORQ. Prior to the distribution, the study received ethical approval from the university and students were given informed consent to opt-in to completing the survey. No incentive was provided for completing the survey. Only students who passed the course were surveyed. Of the 98 students in the population of students who have completed the course in the semesters surveyed, there were 50 completed PRTG responses. While low (50\%), this was already encouraging given response rates for such optional, anonymous university surveys are often 20-30\% (pre-COVID-19) \citep{Jaffray}. Nine responses were from  female students, (18.3\%) which matches the male/female distribution in the overall population of this course. One student did not indicate gender, and 40 respondents identified as male. A usability feedback survey based on the PDT was used as reported in \citep{Weitl}.  

The CARMA survey used 116 words which was a slight adjustment from the 118 words in the PDT set found in \citep{benedek2002a}. For the ZORQ study, the same set of 55 words as \citep{Barnum2010} and shown in the original article \citep{Benedek} were used. The more reasonably sized list of 55 terms for the user to select allowed the list to fit into a single online question, without any scrolling \citep{Veral}. The PDT words were statistically evaluated and categorized into positive, neutral, and negative categories by \citeauthor*{Veral} (\citeyear{Veral}). With consideration for the potential bias of positive feedback, of the 55 words selected, the same percentage of positive terms from the original PDT (60\%) was kept. Negative and neutral terms each accounted for 20\% of the remaining words selected. In the ZORQ dataset, 15 of the 50 users did not provide explanations for the terms selected.

PDT datasets are often small, such as n=29 in \citep{lim} and n=10 in \citep{imler}. This is likely due to challenges inherent in data collection from users as discussed in \citep{hastings2024}. Small data sets are acceptable for user testing experiments, as Alroobaea and Mayhew (\citeyear{Alroobaea}) found that the optimum sample size of 16±4 users provide much validity. The ZORQ dataset had 50 valid PRTG responses and CARMA dataset had 56 valid
PRTG responses.

Prior to analysis, the raw survey data were processed to create a more concise dataset in which each row consisted of a word choice and its corresponding explanation. The data were cleaned to remove any inconsistencies or errors. This included checking for and addressing issues such as missing values, mismatched quotes, and non-text characters.

\subsection{Data Labeling}
To establish a gold standard for evaluation, a human-in-the-loop (HITL) annotation approach \citep{yu2016} was used, with the authors and a research student serving as annotators to perform manual data labeling. 
Although many studies use LLMs such as GPT4 for data labeling \citep{He}, a baseline of manually labeled test data is necessary to understand the accuracy of LLM-based labeling.   In-house HITL manual annotation secures the highest quality labeling possible and is generally considered the gold standard by data scientists and engineers \citep{labelData}. It is a common method, especially for small data sets that require expertise that cannot be easily crowd-sourced \citep{labelData} and to provide a baseline of labeled test data \citep{He, wang2023chatgpt, Mukhin}. \citeauthor*{He} \citet{He} found that GPT4 itself had higher accuracy than crowd-sourcing (particularly Amazon Mechanical Turk (MTurk) workers' pipeline \citep{Murk}, highlighting the 
unreliable quality in crowd-sourced labels; further supporting the necessity for manual in-house labeling in this study. 

The annotators, with over 60 years of combined software development experience, have the necessary expertise to manually annotated the dataset to create a gold-standard to measure the LLMs.  
The first two annotators have PhDs in Computer Science (CS), and have over 30 years experience in software development each. The third annotator is an upper-level undergraduate CS student, with a couple of years of in-school software development experience. Additionally, since the data collected was from undergraduate CS students located in the same state as the third author, he has a good understanding of the language, vocabulary, relevance, sentiment, and context that the students would use in their descriptions, including slang words.

A potential risk of bias occurs when HITL gold-standard annotation is conducted by the researchers. This research reduced this bias by conducting independent annotation, and by including an annotator from a similar user group as the products being investigated.  Each annotator completed the work independently, without reviewing any scores from the other annotators, or from any automated tool. 

The annotation was completed in two ways: 1) by assigning an overall sentiment score for each user's PRTG, and 2) by scoring each individual term and explanation pair. After the independent annotation was completed, the scores were averaged to create the HITL gold standard. The inter-annotator agreement (Pearson's coefficients) between the three annotators for the ZORQ PRTGs was (0.92, 0.96, and 0.96)  and (0.88, 0.89, 0.89) for the word/explanation pairs. CARMA inter-annotator agreement was similar.

The authors expect that in general, product usability sentiment may tend to be 
skewed either negatively or positively, 
rather than following a normal distribution. Based on the manual data labeling, the ZORQ PDT dataset had an average overall gold-standard sentiment rating (on a scale from 0-1)
of 0.76, for both the set of 
PRTGs and the word/explanation pairs, with standard deviations of 0.26 for the word/explanation pairs, and 0.19 for the PRTGs; showing that the data are skewed in the positive direction, and not normally distributed. 

\begin{figure}
\centerline{\includegraphics[width=1.0\columnwidth]
{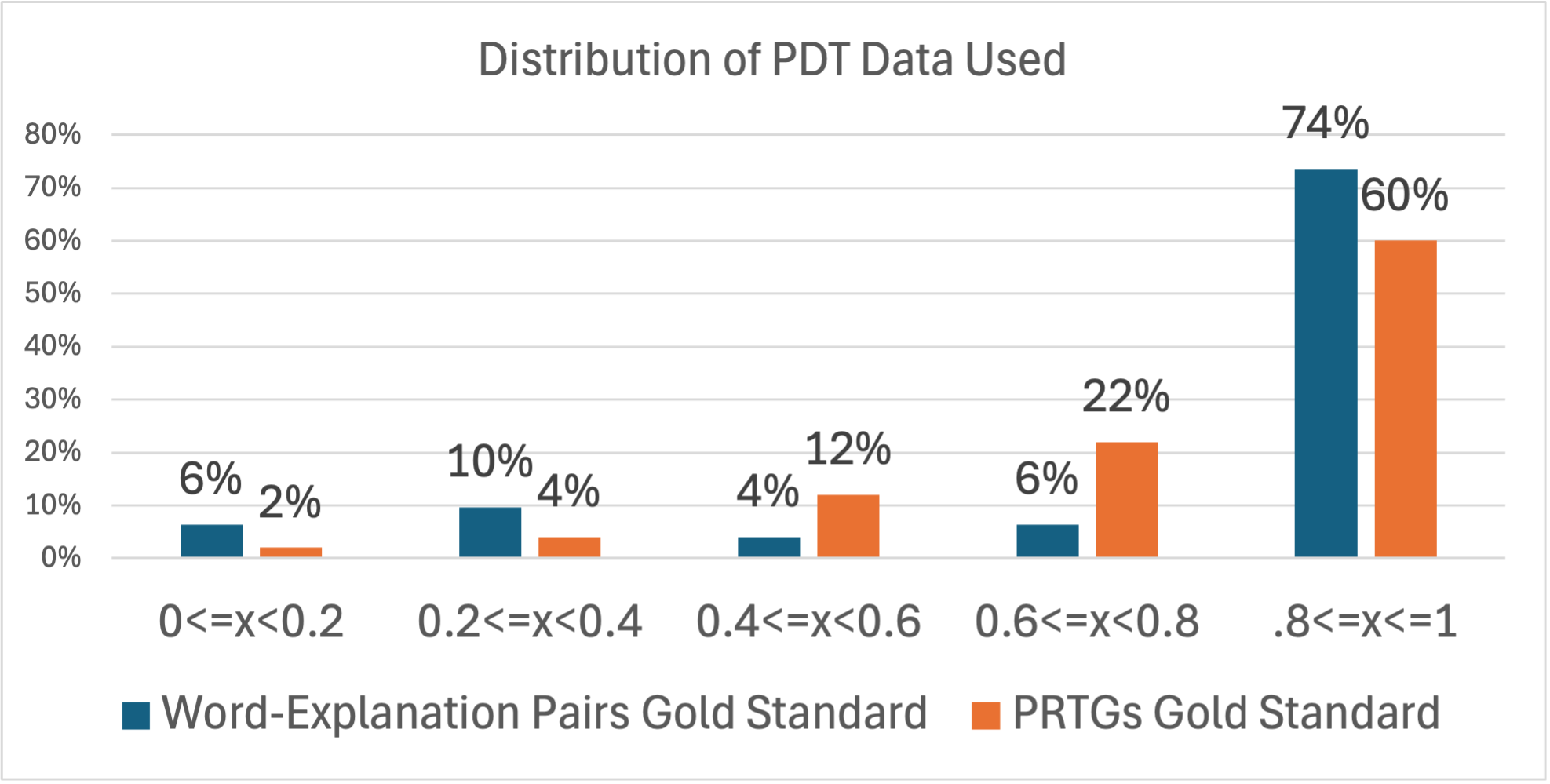}}
 \caption{ZORQ PDT Data Sentiment Distribution}
 \label{DataDistribution}
 \end{figure}

The data were also divided into three classes: Positive (above 0.65), Neutral (between 0.35 and 0.65), and Negative (below 0.35); and into five classes: Very Positive (0.80 and above), Positive (0.6 to 0.8), Neutral (0.4 to 0.6), Negative (0.2 to 0.4), and Very Negative (0.2 and below). The distribution of the ZORQ data grouped into five categories, representing strongly negative to strongly positive, is shown in Figure \ref{DataDistribution}. The ZORQ PDT PRTG dataset had 30 very positive, 11 positive, 6 neutral, 1 negative, and 2 very negative responses. For the word/explanation pairs, the ZORQ data had 184 very positive, 16 positive, 10 neutral, 23 negative, and 17 very negative pairs. When distributed into three categories, the PRTG data had 38 positive, 9 neutral, and 3 negative responses, and the word/explanation pairs had 194 positive, 22 neutral, and 34 negative responses.

\begin{figure}
\centerline{\includegraphics[width=1.0\columnwidth]
{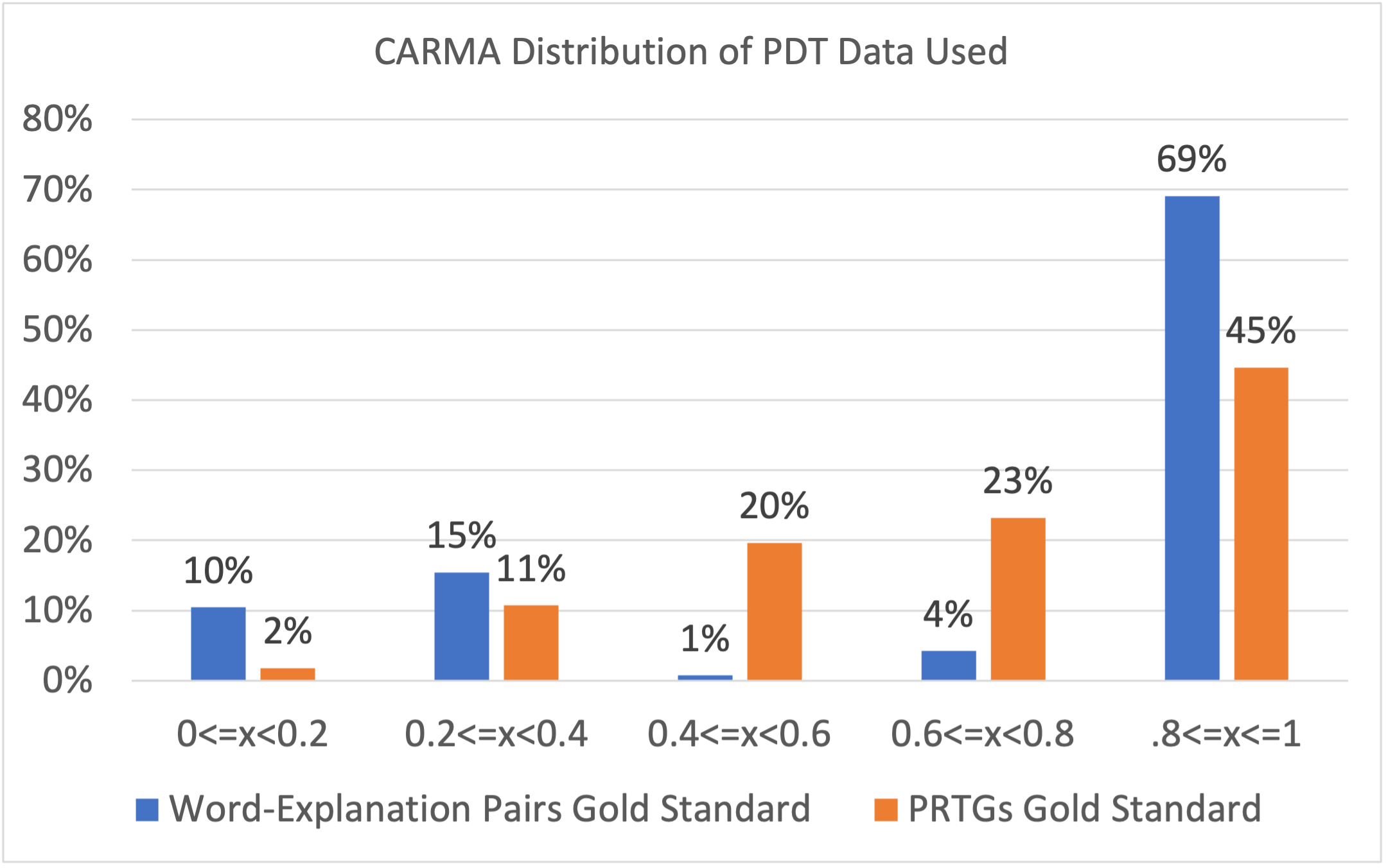}}
 \caption{CARMA PDT Data Sentiment Distribution}
 \label{CarmaDataDistribution}
 \end{figure}

The CARMA data set had an average overall gold-standard sentiment rating for its PRTGs of 0.68 and a standard deviation of 0.23 and for the word/explanation pairs, an average of 0.70 and standard deviation of 0.30. The CARMA PDT PRTG dataset had 25 very positive, 13 positive, 11 neutral, 5 negative, and 2 very negative responses, as shown in Figure \ref{CarmaDataDistribution}. For the word/explanation pairs, the CARMA data had 179 very positive, 10 positive, 2 neutral, 26 negative, and 41 very negative pairs. When distributed into three categories, the PRTG data had 33 positive, 16 neutral, and 6 negative responses, and the word/explanation pairs had 187 positive, 5 neutral, and 66 negative responses.

\subsection{Sentiment Analysis Tools and Methods} 

This research explores sentiment analysis performance on PDT data of several current LLMs, using standard API calls: GPT4 \citep{GPT4}, GPT4o \citep{GPT4o}, GPT4o-mini~\citep{gpt4omini}, Claude Sonnet 3 and 3.5 \citep{Claude}, along with Twitter-RoBERTa-Base-Sen\-timent (TRBS) \citep{Roberta}, and VADER (Valence Aware Dictionary and sEntiment Reasoner) \citep{Hutto_Gilbert_2014}. 
The latter two were chosen for their established effectiveness in sentiment analysis tasks and their differing approaches, which provide a useful comparison to the modern LLMs.  TRBS is a pre-trained bidirectional model fine-tuned on Twitter data for sentiment analysis, based on the RoBERTa architecture, and is a leading transfer learning technique. VADER, a long-standing tool in this domain, combines a robust lexicon with heuristic rules for contextual nuance, making it user-friendly and highly accurate. It was designed to perform well on social media text but is effective across other text forms. 
The selected LLMs (GPT4o and Claude 3.5) represent two advanced LLMs \citep{Krugmann}, and our model is adaptable for use with other LLMs. The GPT4o-mini algorithm was selected to represent a more cost-effective small LLM. LLAMA and Mistral were not used, as ChatGPT was found to excel in linguistic accuracy, with Llama and Mistral considered as excessive and computationally expensive for simple sentiment tasks \citep{Llama}.

Two approaches were attempted for running PDT data through the tools to produce PRTG-level scores: 1) having the tool produce numeric sentiment scores (0-1) for each of the five word/explanation pairs for a respondent and then manually calculating an average (referred to as `Avg5' tests), and 2) having the tool produce one overall sentiment score for all of the word/explanation pairs for a respondent (referred to as `Respondent' tests). From the potential combinations \textit{(tool} \textbf{X} \textit{scoring approach)}, we ran the following tests:
\begin{enumerate}
    \item Claude3-Avg5
    \item Claude3.5-Respondent
    \item GPT4-Avg5
    \item GPT4-Respondent
    \item GPT4o-Avg5
    \item GPT4o-Respondent
    \item GPT4o-mini-Avg5
    \item GPT4o-mini-Respondent
    \item TRBS-Avg5
    \item TRBS-Respondent
    \item VADER-Avg5
    \item VADER-Respondent
\end{enumerate}

For the LLMs, the associated prompts appear in Table \ref{tab:llm_prompts}. The prompts instruct the LLM to score the PDT data and provide an explanation. Except for Claude3, all LLMs were also asked to provide a confidence level (low, medium, high) for their scoring. These prompts were initially developed and refined through the web interfaces for Claude and ChatGPT. The tests for Claude3-Avg5 and Claude3.5-Respondent were run through the web interface by uploading a CSV file of word/explanation pairs to the web interface, while the other prompts were processed through the GPT API. For GPT4o-Avg5 and GPT4o-Respondent, and GPT4o-mini-Avg5 and GPT4o-mini-Respondent, three runs were conducted and the results were averaged. 

\begin{table*}[!htbp]
\centering
\begin{threeparttable}
\caption{LLM Prompts}
\label{tab:llm_prompts}
\begin{tabular}{|p{.95\textwidth}|}
\hline
\textbf{Claude3-Avg5 (Test1):} \\ A group of people were given a survey to assess an experience. Each respondent provided one word to describe their experience and an optional explanation for their choice. The attached spreadsheet contains the responses, one word response per row along with the explanation. In CSV form, provide a sentiment score between 0.00-1.00 (inclusive) for each of the chosen words based on your understanding of the word (with 1.00 being the most positive), along with an adjusted sentiment score for the word based on the context provided by the corresponding explanation. If no explanation is given for a word, the base and adjusted scores should be the same. Each row of output should contain the word, the base score, the adjusted score, and the explanation. Sentiment scores should be to two decimal places. It is very important that you output the results for all 250 responses. \\ \hline

\textbf{Claude3.5-respondent (Test2):} \\ The attached file contains survey responses from 50 respondents. Each respondent selected up to five words to describe their experience, and for each word, optionally provided an explanation for the choice. Please produce an overall sentiment score to two decimal places between 0.00-1.00 (inclusive) for each respondent based on your understanding of language. For each sentiment score, please also provide your confidence in the score (low, medium, high) and a detailed contextual explanation for the sentiment score that a non-technical person can understand. \\ \hline

\textbf{Avg5 (Test 3 GPT4, Test 5 GPT4o \& Test 7 GPT4o-mini) :} \\ The following lines contain word choices, which are sometimes followed by explanations for the choice. For these lines, provide a sentiment analysis score for each word between 0.00-1.00 (to two decimal places), and then an adjusted score for the word based on the explanation. Include your confidence in the accuracy of that score (low, medium, high). Additionally, provide a carefully crafted contextual explanation for the sentiment score that is related to the meaning of the text. Please provide your response in a text-based csv format, with columns for the word, original score, adjusted score, confidence, and explanation. Please do not provide any other response aside from the csv formatted data. Only provide one response per line: \\ \hline

\textbf{Respondent (Test 4 GPT4, Test 6 GPT4o \& Test 8 GPT4o-mini):} \\ Each of the following lines contains a word choice, sometimes followed by an explanation for the choice. Based on this data, please provide a singular sentiment analysis score for all of the words between 0.00-1.00 (to two decimal places), and then a singular adjusted score for all of the words based on all of the explanations. In the event that explanations are not provided with the word choices, please make the adjusted score the same as the original. Include your confidence in the accuracy of that score (low, medium, high). Additionally, provide a carefully crafted contextual explanation for the sentiment score that is related to the meaning of the texts. Please provide your response in a text-based csv format, with columns for the original score, adjusted score, confidence, and explanation. Please do not provide any other response aside from the csv formatted data. Please do not evaluate each line individually, evaluate all of the lines as a whole: \\ \hline

\end{tabular}
\end{threeparttable}
\end{table*}

For Avg5 scoring with TRBS (Test 9) and VADER (Test 11), words and explanations were scored separately and averaged. For respondent-level scoring with TRBS (Test 10) and VADER (Test 12), the word/explanation pairs were concatenated with a separating period and space ``. '', and the five pairs for the respondent were joined in the same way before being processed by the tool.

In addition to numeric sentiment analysis, we also classified the data two different ways, first into three classes: Positive (above 0.65), Neutral (between 0.35 and 0.65), and Negative (below 0.35); and then into five classes: Very Positive (0.80 and above), Positive (0.6 to 0.8), Neutral (0.4 to 0.6), Negative (0.2 to 0.4), and Very Negative (0.2 and below).

We also wanted to see how far we could push GPT4o-mini as compared with the GPT4o LLM in its ability to perform sentiment analysis, and evaluated both GPT4o and GPT4o-mini by withholding various amounts of data, such as only providing the words, and providing the grouped words only for both numerical and classification sentiment analysis.

\subsection{Method Evaluation}

To assess each tool's performance on numerical sentiment analysis, well-established metrics for numerical data analysis, such as Pearson Coefficients (PC), Mean Squared Difference (MSD) and Mean Absolute Difference (MAD) are employed, as done in \citep{weitl2024analyzing}. These metrics provide a comprehensive analysis of sentiment classification, allowing for a thorough evaluation of the system's effectiveness \citep{Bharadwaj}. The two-tailed paired t-test  with $\alpha=0.05$ was used with the null hypothesis that the mean difference between algorithmic results and the gold standard is zero. Additionally, the Wilcoxon statistical test \citep{Rey2011} was used to verify that the data fit the requirements necessary for using the paired t-test.  The Wilcoxon test is particularly useful when dealing small paired sets of data which do not follow a normal distribution. 

Similar to \citep{hastings2024,gelman2025scalable}, GPT4o and GPT4o-mini are evaluated for cost and time efficiencies, based on token costs identified by \citep{openAIpricing}.

For the classification of categorical data, the weighted-average F1, the macro-average F1, and accuracy were used for method evaluation, similar to \citep{Abdelsamie}.

We lastly look at the confidence ratings and explanations provided by the LLMs for understanding their value to implicit user sentiment analysis. When prompted to do so, LLMs, as reasoning machines, have the potential to provide explanations for both qualitative and quantitative sentiment classification tasks, 
bridging the explainability gap between the classification tasks \citep{Krugmann}.

\section{Results} \label{results}
\subsection{Numerical Sentiment Analysis Results (RQ1,2)}

\begin{figure*}
\centerline{\includegraphics[width=1.0\linewidth]
{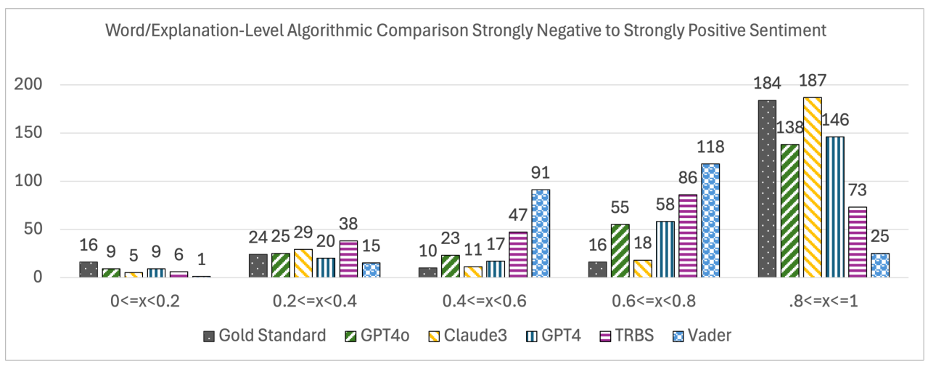}}
 \caption{ZORQ PRTG  Sentiment Distribution}
 \label{PRTGResults}
\end{figure*}

Results showing the numerical sentiment of the ZORQ data divided into five equal ranges from 0 to 1 are shown in Figure \ref{PRTGResults} and Table \ref{tab:sentimentresultsBins}, where tools are listed by decreasing order of Pearson Coefficient (PC). Statistical analysis of the ZORQ results are shown in Table \ref{tab:sentimentresultsStats}, also in decreasing order of PC.
All strong values are highlighted, and all LLMs produced statistically significant $\alpha=0.05$ results at the respondent-level. The GPT4o-Respondent approach (Test 6) produced the best statistically significant results, closely followed by GPT4o-mini-Respondent (Test 8), and GPT4-Respondent (Test 4).  The other LLM at the respondent level (Claude3.5 Test 2) was statistically significant with $\alpha=0.10$.
For the LLM Avg5 tests,  GPT4o-Avg5 (Test 5) and GPT4o-mini-Avg5 (Test 7) results had strong MAD, MSD, and PC results, but were not statistically matched at $\alpha=0.05$, while the Claude3-Avg5 was statistically matched. 
The more traditional approaches, TRBS and VADER performed the poorest under both evaluations, and were not statistically significant.

\definecolor{lgray}{gray}{0.9}
\definecolor{lgreen}{rgb}{0.56, 0.93, 0.56} 

\begin{table*}[ht]
\centering
\begin{threeparttable}
\caption{Comparison of ZORQ PRTG-level Sentiment Analysis Results Ordered by PC}
\label{tab:sentimentresultsBins}
\begin{tabular}{|c|c|c|c|c|c|c|}
\hline
\textbf{Model/Test} & \textbf{$0<=x<0.2$} & \textbf{$0.2<=x<0.4$} & \textbf{$0.4<=x<0.6$} & \textbf{$0.6<=x<0.8$} & \textbf{$0.8<=x<=1$} & \textbf{Totals} \\ \hline
\cellcolor{lgray}\textbf{Gold Standard} & \cellcolor{lgray}1 & \cellcolor{lgray}2 & \cellcolor{lgray}6 & \cellcolor{lgray}11 & \cellcolor{lgray}30 & \cellcolor{lgray}50 \\ \hline
\textbf{Test(6) 3Run$\bar{x}$} & 2 & 1 & 5 & 14 & 28 & 50 \\ \hline
\textbf{Test(1)} & 1 & 1 & 3 & 15 & 30 & 50 \\ \hline
\textbf{Test(5) 3Run$\bar{x}$} & 1 & 2 & 6 & 24 & 17 & 50 \\ \hline
\textbf{Test(2)} & 1 & 1 & 2 & 13 & 33 & 50 \\ \hline
\textbf{Test(7) 3Run$\bar{x}$} & 1 & 1 & 5 & 25 & 18 & 50 \\ \hline
\cellcolor{lgreen}\textbf{Test(8) 3Run$\bar{x}$} & 2 & 1 & 9 & 13 & 25 & 50 \\ \hline
\textbf{Test(4)} & 2 & 1 & 4 & 11 & 32 & 50 \\ \hline
\textbf{Test(3)} & 0 & 2 & 5 & 23 & 20 & 50 \\ \hline
\textbf{Test(10)} & 3 & 0 & 4 & 9 & 34 & 50 \\ \hline
\textbf{Test(12)} & 2 & 0 & 1 & 7 & 40 & 50 \\ \hline
\textbf{Test(9)} & 0 & 0 & 8 & 37 & 5 & 50 \\ \hline
\textbf{Test(11)} & 0 & 0 & 35 & 15 & 0 & 50 \\ \hline
\end{tabular}
\end{threeparttable}
\end{table*}

\begin{table*}[ht]
\centering
\begin{threeparttable}
\caption{Comparison of ZORQ PRTG-level Sentiment Analysis Results Ordered by PC}
\label{tab:sentimentresultsStats}
\begin{tabular}{|c|c|c|c|c|c|c|c|c|}
\hline
\textbf{Model/Test} & \textbf{MAD} & \textbf{MSD} & \textbf{Mn} & \textbf{Mx} & \textbf{Avg} & \textbf{SD} & \textbf{PC} & \textbf{t Stat} \\ \hline
\cellcolor{lgray}\textbf{Gold Standard} &  &  & \cellcolor{lgray}0.10 & \cellcolor{lgray}0.94 & \cellcolor{lgray}0.76 & \cellcolor{lgray}0.19 &  &  \\ \hline
\textbf{Test(6) 3Run$\bar{x}$} & \cellcolor{lgreen}0.04 & \cellcolor{lgreen}0.00 & 0.09 & 0.97 & \cellcolor{lgreen}0.77 & 0.20 & \cellcolor{lgreen}0.97 & \cellcolor{lgreen}-1.29 \\ \hline
\textbf{Test(1)} & \cellcolor{lgreen}0.03 & \cellcolor{lgreen}0.00 & 0.10 & 0.94 & \cellcolor{lgreen}0.77 & 0.17 & \cellcolor{lgreen}0.97 & \cellcolor{lgreen}-1.78 \\ \hline
\textbf{Test(5) 3Run$\bar{x}$} & \cellcolor{lgreen}0.06 & \cellcolor{lgreen}0.00 & 0.15 & 0.87 & \cellcolor{lgreen}0.72 & 0.15 & \cellcolor{lgreen}0.97 & 5.95 \\ \hline
\textbf{Test(2)} & \cellcolor{lgreen}0.05 & \cellcolor{lgreen}0.00 & 0.15 & 0.95 & \cellcolor{lgreen}0.79 & 0.18 & \cellcolor{lgreen}0.95 & -3.27 \\ \hline
\textbf{Test(7) 3Run$\bar{x}$} & \cellcolor{lgreen}0.07 & \cellcolor{lgreen}0.00 & 0.03 & 0.89 & \cellcolor{lgreen}0.74 & 0.15 & \cellcolor{lgreen}0.94 & 3.99 \\ \hline
\cellcolor{lgreen}\textbf{Test(8) 3Run$\bar{x}$} & \cellcolor{lgreen}0.07 & \cellcolor{lgreen}0.01 & 0.15 & 1.00 & \cellcolor{lgreen}0.76 & 0.23 & \cellcolor{lgreen}0.93 & \cellcolor{lgreen}-1.69 \\ \hline
\textbf{Test(4)} & \cellcolor{lgreen}0.06 & \cellcolor{lgreen}0.01 & 0.05 & 1.00 & \cellcolor{lgreen}0.78 & 0.20 & \cellcolor{lgreen}0.92 & \cellcolor{lgreen}-1.62 \\ \hline
\textbf{Test(3)} & \cellcolor{lgreen}0.07 & \cellcolor{lgreen}0.01 & 0.31 & 0.88 & \cellcolor{lgreen}0.72 & 0.13 & \cellcolor{lgreen}0.89 & -2.01 \\ \hline
\textbf{Test(10)} & 0.12 & \cellcolor{lgreen}0.02 & 0.04 & 0.99 & 0.82 & 0.24 & 0.84 & 3.28 \\ \hline
\textbf{Test(12)} & 0.14 & \cellcolor{lgreen}0.03 & 0.03 & 1.00 & 0.88 & 0.21 & 0.78 & 6.47 \\ \hline
\textbf{Test(9)} & 0.16 & \cellcolor{lgreen}0.04 & 0.54 & 0.91 & 0.68 & 0.08 & 0.11 & 2.80 \\ \hline
\textbf{Test(11)} & 0.23 & 0.07 & 0.48 & 0.74 & 0.58 & 0.07 & -0.05 & 6.18 \\ \hline
\end{tabular}
\end{threeparttable}
\end{table*}

While not shown in the tables, the three GPT4o respondent-level independent runs (Test 6) on the ZORQ data were statistically matched, and had between-run PC values of 0.90, 0.92, and 0.92. The three GPT4o-mini respondent-level runs (Test 8) were also statistically matched and had strong between-run PCs (0.93). For the three GPT4o Avg5 runs (Test 5), the t-stats for run\textsubscript{1} and run\textsubscript{2} means were not statistically matched, but the other runs had between-run t-stats that were significant, with between-run PC values of 0.85, 0.82, and 0.95. GPT4o-mini Avg5 runs (Test 7) were statistically matched, and had strong between-run PC values (0.93).

\begin{table*}[ht]
\centering
\begin{threeparttable}
\caption{ Comparison of CARMA PRTG-level Sentiment Analysis Results Ordered by PC}
\label{tab:CARMAsentimentresultsStats}
\begin{tabular}{|c|c|c|c|c|c|c|c|c|}
\hline
\textbf{Model/Test} & \textbf{MAD} & \textbf{MSD} & \textbf{Mn} & \textbf{Mx} & \textbf{Avg} & \textbf{SD} & \textbf{PC} & \textbf{t Stat} \\ \hline
\cellcolor{lgray}\textbf{Gold Standard} &  &  & \cellcolor{lgray}0.13 & \cellcolor{lgray}0.93 & \cellcolor{lgray}0.68 & \cellcolor{lgray}0.23 &  &  \\ \hline
\textbf{Test(1) }&\cellcolor{lgreen}0.06&\cellcolor{lgreen}0.01&0.26&0.90&\cellcolor{lgreen}0.70&0.17&\cellcolor{lgreen}0.97&\cellcolor{lgreen}-1.48\\ \hline
\textbf{Test(3) }&\cellcolor{lgreen}0.05&\cellcolor{lgreen}0.00&0.13&0.94&\cellcolor{lgreen}0.67&0.23&\cellcolor{lgreen}0.97&\cellcolor{lgreen}0.85\\ \hline
\textbf{Test(5) 3Run$\bar{x}$} & \cellcolor{lgreen}0.06 & \cellcolor{lgreen}0.01 & 0.08 & 0.94 & \cellcolor{lgreen}0.66 & 0.27 & \cellcolor{lgreen}0.96 & 6.25 \\ \hline
\textbf{Test(7) 3Run$\bar{x}$} & \cellcolor{lgreen}0.07 & \cellcolor{lgreen}0.01 & 0.05 & 0.95 & \cellcolor{lgreen}0.66 & 0.28 & \cellcolor{lgreen}0.96 & 5.98 \\ \hline
\textbf{Test(2)}&\cellcolor{lgreen}0.05&\cellcolor{lgreen}0.01&0.20&0.95&\cellcolor{lgreen}0.67&0.22&\cellcolor{lgreen}0.95&\cellcolor{lgreen}1.37\\ \hline
\cellcolor{lgreen}\textbf{Test(6) 3Run$\bar{x}$} & \cellcolor{lgreen}0.07 & \cellcolor{lgreen}0.01 & 0.14 & 0.96 & \cellcolor{lgreen}0.68 & 0.24 & \cellcolor{lgreen}0.94 & \cellcolor{lgreen}0.06 \\ \hline
\textbf{Test(8) 3Run$\bar{x}$} & \cellcolor{lgreen}0.09 & \cellcolor{lgreen}0.01 & 0.00 & 1.00 & \cellcolor{lgreen}0.64 & 0.29 & \cellcolor{lgreen}0.93 & 3.05 \\ \hline
\textbf{Test(10)}&0.13&\cellcolor{lgreen}0.03&0.03&0.98&0.62&0.32&0.89&2.79\\ \hline
\textbf{Test(9) }&0.16&\cellcolor{lgreen}0.04&0.18&0.90&0.55&0.14&0.85&7.39\\ \hline
\textbf{Test(4) }&\cellcolor{lgreen}0.07&\cellcolor{lgreen}0.01&0.22&0.90&\cellcolor{lgreen}0.70&0.19&0.76&-1.48\\ \hline
\textbf{Test(11) }&0.16&\cellcolor{lgreen}0.05&0.06&0.99&0.76&0.28&0.71&-2.90\\ \hline
\textbf{Test(12) }5&0.19&\cellcolor{lgreen}0.05&0.39&0.79&0.56&0.08&0.70&4.96\\ \hline
\end{tabular}
\end{threeparttable}
\end{table*}

For the respondent-level tests on the CARMA dataset, GPT4o, GPT4, and Claude3.5 all  produced statistically significant results, while GPT4o-mini was just outside of statistical significance, but had strong (0.93) PC, (0.04) MAD and (0.00) MSD values, as shown in Table \ref{tab:CARMAsentimentresultsStats}. For the LLM Avg5 tests,  GPT4o-Avg5 and GPT4o-mini-Avg5 results had strong MAD, MSD, and PC results, but were not statistically matched, while the Claude3-Avg5 and GPT4 were statistically matched. For CARMA data, TRBS and VADER performed the poorest under both evaluations, and were not statistically significant. All strong values are highlighted. The GPT4o-Respondent approach (Test 6) produced the best statistically significant results.

\subsection{PDT Dataset Costs (RQ3)}

Table \ref{tab:gen_costs} shows the costs of data generation in terms of time, tokens for the GPT4o and GPT4o-mini test runs, and price per method for the ZORQ data. 
The input sizes (i.e., the prompts sent to GPT4o-mini and GPT4o) were relatively similar in size. Since three runs were conducted, the table includes the total for the three runs of each of these tests. Given that the datasets had similar sizes, the test runs had similar costs for the CARMA data set.

\begin{table}[htbp]
\centering
\begin{threeparttable}
\caption{Costs for the ZORQ PDT Datasets}
\label{tab:gen_costs}
\begin{tabular}{|c|c|c|c|c|c|}
  \hline
&  & \multicolumn{3}{c|}{\textbf{Tokens}} & \\ \cline{3-5}
\textbf{Test} & \textbf{Time (s)} & \textbf{Input} & \textbf{Output} & \textbf{Total} & \textbf{Price} \\ \hline
Test(8)& 337.93 & 46074 & 13396 & 59470 & \$0.75 \\ \hline
Test(6) & 277.14 & 46074 & 14013 & 60087 & \$12.50 \\ \hline
Test(7) & 463.88 & 22659  & 27784 & 50443 & \$0.75 \\ \hline
Test(5) & 523.70 & 22659  & 23716 & 46375 & \$12.50 \\ \hline
\end{tabular}
\end{threeparttable}
\end{table}

The times required to run the tests were relatively low, averaging 400 seconds for three runs of each test. Notice, the time to run for the GPT4o-mini tests (Tests 7 and 8) was comparable to its GPT4o counterpart (Tests 5 and 6).  

 At the time of running the experiments in Spring 2025, GPT4o-mini dollar costs were relatively low at \$0.15 per 1M input tokens and \$0.60 per 1M output tokens~\citep{openAIpricing}, 
making it more ideal for big data synthesis tasks compared to some of the other massive commercial LLMs. For example, GPT4o-mini represents a 94\% savings over GPT4o which costs \$2.50 per 1M input tokens and \$10.00 per 1M output tokens~\citep{openAIpricing}. For the experiments conducted in this research, the GPT4o-mini cost was \$0.75 for both the respondent-level and Avg-5 tests, while the GPT4o costs were \$12.50. 
Scaling up generation, the estimated costs to analyze a PDT dataset with 1M responses using term/explanation (the cheapest approach) would be $\sim$\$402 with GPT4o-mini, and $\sim$\$5,877 with GPT4o.

\subsection{Categorical Sentiment Evaluation (RQ4)}

In addition to numeric sentiment analysis, we experimented with GPT4o and GPT4o-mini  classifying the sentiment of the PRTGs and the individual word/explanation pairs two different ways, first into three classes: Positive (above 0.65), Neutral (between 0.35 and 0.65), and Negative (below 0.35); and then into five classes: Very Positive (0.80 and above), Positive (0.6 to 0.8), Neutral (0.4 to 0.6), Negative (0.2 to 0.4), and Very Negative (0.2 and below).

Using five categories, the ZORQ data had 30 (60\%) very positive PRTGs, 11 (22\%) positive, 6 (12\%) neutral, 1 negative (2\%) and and 2 (4\%) very negative PRTGs. When using five categories, the accuracy and F1-measure values were not as good as the values for three categories for the ZORQ data, as shown in Tables \ref{tab:model_comparison_4a} and \ref{tab:model_comparison_4b}. There does not appear to be a clear advantage to using GPT4o over GPT4o-mini for categorical classification, as both LLMs sometimes outperform the other, depending on the way the test was run. In reviewing the confusion matrices for the PRTG data such as in Table \ref{tab:confusionMatrix_4a}, GPT4o-mini labeled 24 PRTG sets correctly as ``very positive", but missed 6 by labeling them as ``positive", and mislabeled 20 of the 50 PRTGs.  GPT4o made similar mistakes ``very positive" and ``positive". This may be caused by inappropriate cutoff values for the data set. 

Using three categories, the ZORQ data had 38 (76\%) positive PRTGs, 9 (18\%) neutral PRTGs, and 3 (6\%) negative PRTGs. The accuracy and F1-values for the three categories were acceptable, with accuracy ranging from 88\%-94\% as shown in Table \ref{tab:model_comparison_4b}. The confusion matrix in Table \ref{tab:confusionMatrix_4b} shows much fewer false results (5 out of 50) using three categories as compared with five categories.

\begin{table*}
\centering
\begin{threeparttable}
\caption{Categorical sentiment analysis for the ZORQ PDT dataset (five categories)}
\label{tab:model_comparison_4a}
\setlength{\tabcolsep}{6pt} 
\begin{tabular}{|l|c|c|c|c|}
\hline
 & mini-PRTG & 4o-PRTG & mini word/expl & 4o word/expl \\ \hline
\textbf{accuracy} & 76.00 & 67.60 & 63.60 & 86.00 \\ \hline
\textbf{macro-averaged F1} & 0.62 & 0.51 & 0.47 & 0.89 \\ \hline
\textbf{weighted-averaged F1} & 0.80 & 0.76 & 0.73 & 0.87 \\ \hline
\end{tabular}
\end{threeparttable}
\end{table*}

\begin{table*}
\centering
\begin{threeparttable}
\caption{Confusion Matrix for the ZORQ PDT dataset (five categories, mini-PRTG)}
\label{tab:confusionMatrix_4a}
\small
\begin{tabular}{|c|c|c|c|c|c|c|}
\hline
&Actual: Very Pos&Pos&Neutral&Neg&Very Neg&True Pos+ False Pos \\ \hline
Pred: Very Pos&24&1&0&0&0&25 \\ \hline
Pred: Pos&6&7&0&0&0&13 \\ \hline
Pred: Neutral&0&3&5&1&0&9 \\ \hline
Pred: Neg&0&0&1&0&0&1 \\ \hline
Pred: Very Neg&0&0&0&0&2&2 \\ \hline
True Pos + False Neg&30&11&6&1&2& \\ \hline
\end{tabular}
\end{threeparttable}
\end{table*}

\begin{table*}[h]
\centering
\begin{threeparttable}
\caption{Categorical sentiment analysis for the ZORQ PDT dataset (three categories)}
\label{tab:model_comparison_4b}
\setlength{\tabcolsep}{6pt}
\begin{tabular}{|l|c|c|c|c|}
\hline
 & mini-PRTG & 4o-PRTG & mini word/expl & 4o word/expl \\ \hline
\textbf{accuracy} & 88.00 & 94.00 & 89.60 & 91.60 \\ \hline
\textbf{macro-averaged F1} & 0.77 & 0.87 & 0.74 & 0.79 \\ \hline
\textbf{weighted-averaged F1} & 0.89 & 0.94 & 0.92 & 0.93 \\ \hline
\end{tabular}
\end{threeparttable}
\end{table*}

\begin{table*}
\centering
\begin{threeparttable}
\caption{Confusion Matrix for the ZORQ PDT dataset (three categories, mini-PRTG)}
\label{tab:confusionMatrix_4b}
\small
\begin{tabular}{|c|c|c|c|c|}
\hline
&Actual: Pos&Neutral&Neg&True Pos+ False Pos \\ \hline
Pred: Positive&35&1&0&36\\ \hline
Pred: Neutral&3&7&1&11\\ \hline
Pred: Negative&0&1&2&3\\ \hline
True Positive + False Negative&38&9&3&\\ \hline
\end{tabular}
\end{threeparttable}
\end{table*}

An analysis on the CARMA dataset found similar results, with the three category accuracies ranging from 78\%-94\%, and the mislabeling between strongly positive and  positive values also occurred when using five categories.

\subsection{Efficiency Evaluation (RQ5)}

The comparison of GPT4o-mini to GPT4o in their ability to perform sentiment analysis of PDT data by withholding various amounts of data provided interesting results. 
We compared both the numerical and the classification sentiment analysis provided by the LLMs, using as input, in the following tests:
\begin{itemize}
  \item[A)] the aggregate (respondent-grouped) terms
  \item[B)] the aggregate (respondent-grouped) term + explanation pairs
  \item[C)] the individual terms
  \item[D)] the individual term + explanation pairs
\end{itemize}

\begin{figure*}
\centerline{\includegraphics[width=1.0\linewidth]
{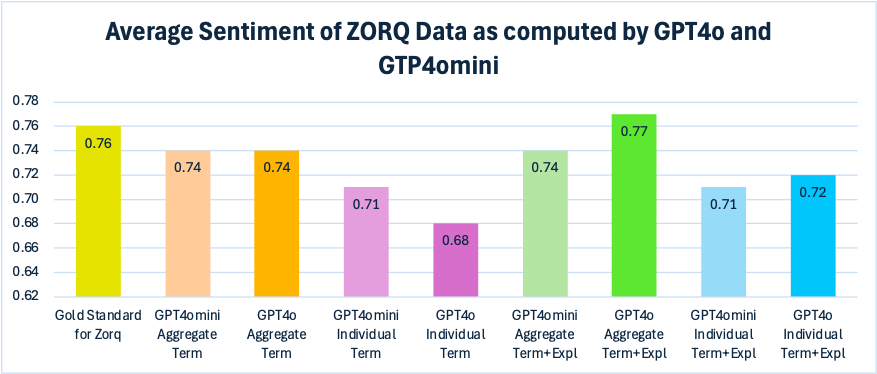}}
\caption{Average Sentiment of ZORQ PDT as Computed by GPT4o and GPT4o-mini}
\label{averageZORQ}
\end{figure*}

As shown in Figure \ref{averageZORQ} and Table \ref{tab:zorq_stats}, 
GPT4o-mini had more accurate scores and had stronger statistical significance than GPT4o for (Test A) aggregate term scoring of the ZORQ data, where the gold standard average sentiment was 0.76. GPT4o-mini and GPT4o had comparable and statistically significant scores for (Test B) aggregate term+explanation scoring (also reported in Table \ref{tab:sentimentresultsStats} as Tests 6 and 8). The LLMs provided statistically significant results for both tests (A) and (B). For Tests (C) and (D), the LLMs had good MAD, MSD, and PC values, but their results were not statistically significant.

\begin{figure*}
\centerline{\includegraphics[width=1.0\linewidth]
{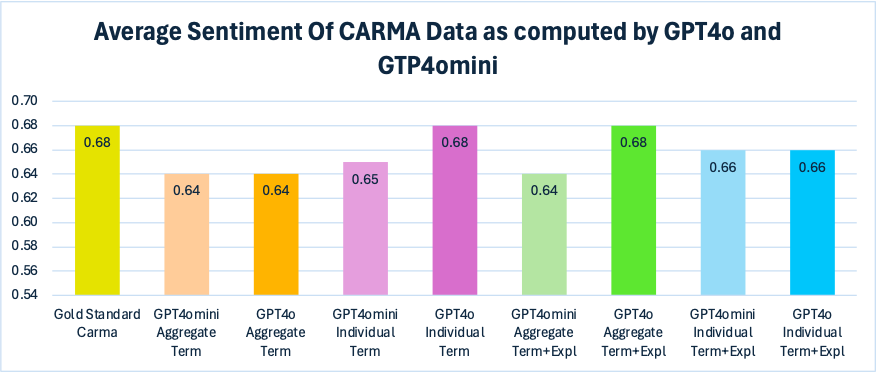}}
\caption{Average Sentiment of CARMA PDT as Computed by GPT4o and GPT4o-mini}
\label{averageCARMA}
\end{figure*}

As shown in Figure \ref{averageCARMA} and Table \ref{tab:carma_stats}, 
GPT4o and GPT4o-mini had similar averages scores, but GPT4o-mini had stronger statistical significance than GPT4o for (Test A) aggregate term scoring of the CARMA data, where the gold standard average sentiment was 0.68. GPT4o had slightly better average and and was statistically significant for (Test B) aggregate term+explanation scoring (also reported in Table \ref{tab:CARMAsentimentresultsStats} as Tests 6 and 8).  GPT4o had a slightly better average and was statistically significant for (Test C) aggregate term+explanation scoring.
For Test D, the LLMs had good MAD, MSD, and PC values, but their results were not statistically significant (also reported in Table \ref{tab:CARMAsentimentresultsStats} as Tests 5 and 7). Also shown in Figure \ref{averageCARMA} and Table \ref{tab:carma_stats},  GPT-4o-mini was slightly worse than GPT-4o in relation to the gold standard for aggregate term+explanation scoring of the CARMA data, where the gold standard average sentiment was 0.68. GPT4o-mini and GPT4o had comparable and statistically significant scores for (Test B) aggregate term+explanation scoring, with 4o-mini having a Pearson (r) = 0.93 and t-Stat = 3.05, and 4o having a Pearson (r) = 0.94 and t-Stat = 0.06.  Other sentiment scores for the CARMA efficiency tests shown in Figure \ref{averageCARMA} followed the positive and negative sentiment trends of the gold standard scoring but were not statistically significant.

\begin{table}
\centering
\begin{threeparttable}
\setlength{\tabcolsep}{4pt}
\caption{ZORQ results under various aggregation and\\ explanation conditions.}
\label{tab:zorq_stats}
\begin{tabular}{|c||c|c|c|c|}
\hline
\multicolumn{5}{|c|}{\textbf{GPT-4o-mini} } \\ \hline
\textbf{ZORQ} & Test A & Test B & Test C& Test D \\
\hline
\textbf{MAD} & 0.07 & 0.07 & 0.07 & 0.09 \\
\hline
\textbf{MSD} & 0.01 & 0.01 & 0.01 & 0.01 \\
\hline
\textbf{Mn} & 0.05 & 0.03 & 0.10 & 0.02 \\
\hline
\textbf{Mx} & 0.97 & 1.00 & 0.92 & 0.97 \\
\hline
\textbf{Avg} & \textbf{0.74} & \textbf{0.74} & \textbf{0.71} & \textbf{0.71} \\
\hline
\textbf{SD} & 0.23 & 0.23 & 0.25 & 0.22 \\
\hline
\textbf{Pearson} & 0.91 & 0.93 & 0.93 & 0.92 \\
\hline
\textbf{t Stat} & 1.13 & 1.69 & 7.21 & 6.40 \\
\hline
\multicolumn{5}{|c|}{\textbf{GPT-4o}} \\
\hline
\textbf{ZORQ} & Test A & Test B & Test C & Test D \\
\hline
\textbf{MAD} & 0.05 & 0.04 & 0.11 & 0.08 \\
\hline
\textbf{MSD} & 0.00 & 0.00 & 0.02 & 0.01 \\
\hline
\textbf{Mn} & 0.20 & 0.12 & 0.10 & 0.12 \\
\hline
\textbf{Mx} & 0.92 & 0.94 & 0.90 & 0.95 \\
\hline
\textbf{Avg} & \textbf{0.74} & \textbf{0.77} & \textbf{0.68} & \textbf{0.72} \\
\hline
\textbf{SD} & 0.18 & 0.19 & 0.22 & 0.22 \\
\hline
\textbf{Pearson} & 0.95 & 0.97 & 0.92 & 0.95 \\
\hline
\textbf{t-stat} & 2.50 & -1.82 & 11.18 & 7.50 \\
\hline
\end{tabular}
\end{threeparttable}
\end{table}

\begin{table}
\centering
\begin{threeparttable}
\setlength{\tabcolsep}{4pt}
\caption{CARMA results under various aggregation and\\ explanation conditions.}
\label{tab:carma_stats}
\begin{tabular}{|c||c|c|c|c|}
\hline
\multicolumn{5}{|c|}{\textbf{GPT-4o-mini} } \\ \hline
\textbf{CARMA} & Test A & Test B & Test C& Test D \\
\hline
\textbf{MAD} & 0.09 & 0.09 & 0.07 & 0.09 \\
\hline
\textbf{MSD} & 0.01 & 0.01 & 0.01 & 0.01 \\
\hline
\textbf{Mn} & 0.00 & 0.00 & 0.08 & 0.01 \\
\hline
\textbf{Mx} & 0.97 & 1.00 & 0.92 & 0.95 \\
\hline
\textbf{Avg} & \textbf{0.64} & \textbf{0.64} & \textbf{0.65} & \textbf{0.66} \\
\hline
\textbf{SD} & 0.30 & 0.29 & 0.30 & 0.28 \\
\hline
\textbf{Pearson} & 0.94 & 0.93 & 0.94 & 0.96 \\
\hline
\textbf{t Stat} & 2.51 & 3.05 & 7.88 & 5.98 \\
\hline
\multicolumn{5}{|c|}{\textbf{GPT-4o}} \\

\hline
\textbf{CARMA} & Test A & Test B & Test C & Test D \\
\hline
\textbf{MAD} & 0.07 & 0.07& 0.11 & 0.06 \\
\hline
\textbf{MSD} & 0.01 & 0.01 & 0.03 & 0.01 \\
\hline
\textbf{Mn} & 0.17 & 0.14 & 0.13 & 0.08 \\
\hline
\textbf{Mx} & 0.89 & 0.96 & 0.91 & 0.94 \\
\hline
\textbf{Avg} & \textbf{0.64} & \textbf{0.68} & \textbf{0.68} & \textbf{0.65} \\
\hline
\textbf{SD} & 0.28 & 0.24 & 0.23 & 0.27 \\
\hline
\textbf{Pearson} & 0.94 & 0.94 & 0.82 & 0.96 \\
\hline
\textbf{t-stat} & 3.64 & 0.06 & 1.14 & 6.25 \\
\hline
\end{tabular}
\end{threeparttable}
\end{table}

Additionally, the 250 ZORQ individual term+explana\-tion pairs (Test D) were evaluated through the other algorithms. The PC values for the algorithms are Claude3  0.93, GPT4 0.85, TRBS 0.80, and VADER 0.64. The paired sample t-Test And Wilcoxon test failed for all of these algorithms. 

On the CARMA data, GPT4o-mini was slightly worse than GPT4o in relation to the Gold Standard for aggregate term+explanation scoring, with GPT4o-mini having a PC of 0.93, and a less statistically significant t-stat of 3.05 than GPT4o's PC of 0.94 and a t-stat = 0.06. Other sentiment scores for CARMA followed the positive and negative sentiment trends of the Gold Standard scoring but were not statistically significant.

\subsection{LLM Confidence (RQ6)}

\begin{figure*}
\centerline{\includegraphics[width=1.0\linewidth]
{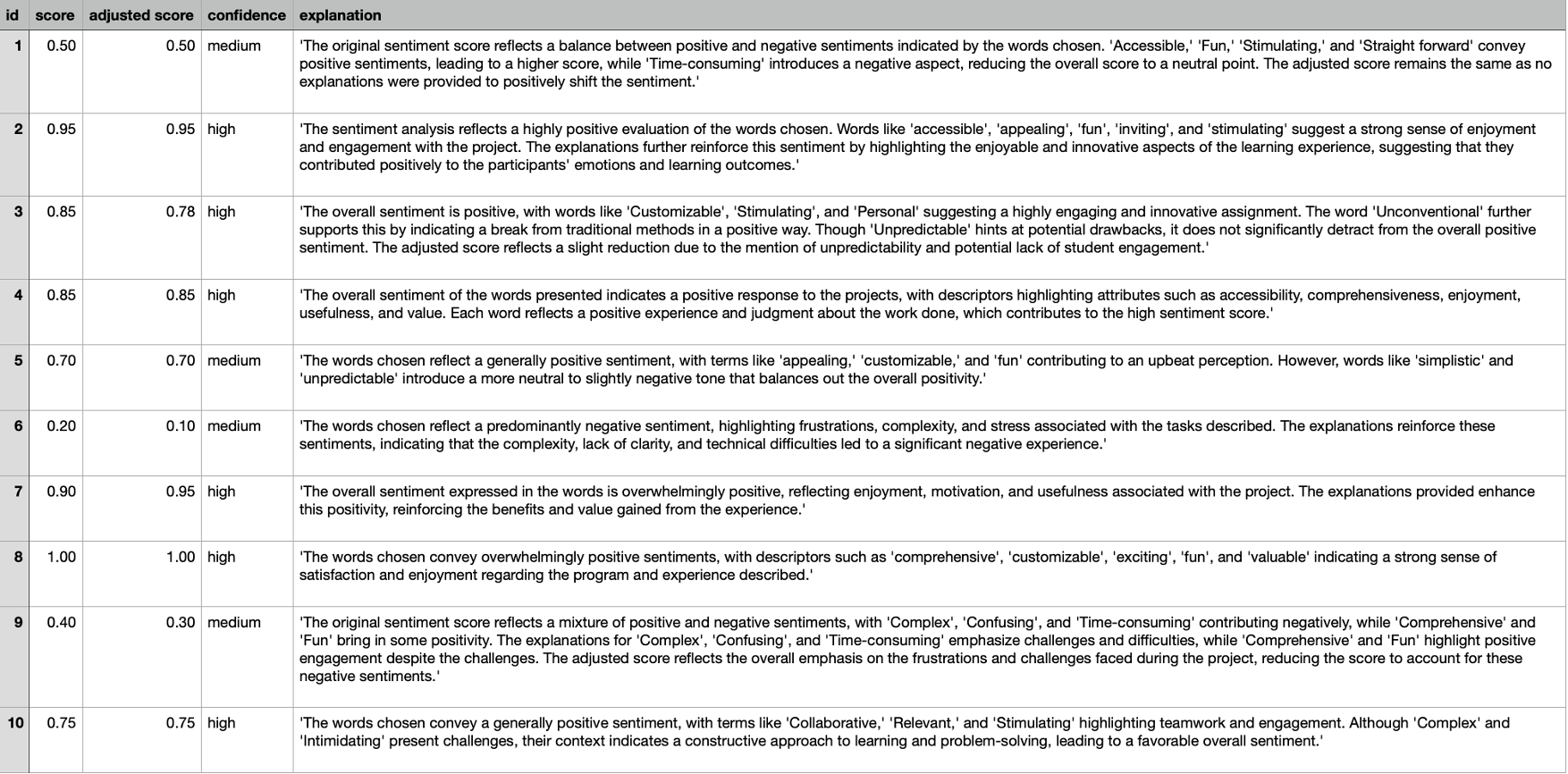}}
 \caption{Sample Output for a Single Run using ZORQ data for GPT-4o-mini, including the Confidence and Explanation}
 \label{output}
 \end{figure*}

All LLMs were asked to provide confidence and explanation on all of their runs. A sample output from GPT4o-mini is shown in Figure \ref{output}. 
All GPT4o runs expressed high confidence for 36 PRTGs and medium for three PRTGs for the ZORQ data. On average, GPT4o expressed high confidence on 42 of the 50 PRTGs, and medium confidence on eight. GPT4o did not express low confidence on any PRTG during any run.  All GPT4o-mini runs expressed high confidence for 30 and medium on 14 PRTGs for the ZORQ data. On average, GPT4o-mini expressed high confidence on 32 of the 50 PRTGs, and medium confidence on 17. GPT4o-mini did not express low confidence on any PRTG during any run. Claude3.5 expressed high confidence on 31, medium on 8, and low on 11 PRTGs. GPT4 expressed high confidence on 42, and medium on eight PRTGs. 

An example where GPT4o (0.82), GPT4o-mini (0.85) and Claude3.5 (0.90) all expressed high confidence for a PRTG with positive sentiment (0.90) was explained by GPT4o as: 
``The sentiment across all of the words is generally positive, reflecting feelings of accomplishment, enjoyment, and educational benefit. The provided explanations offer clarity and reinforcement of this positive sentiment. The terms `Accessible', `Comprehensive', `Fun', `Useful', and `Valuable' are described in contexts that emphasize understanding, thoroughness, enjoyment, and utility, all of which contribute to an overall positive impression.'' (run1).  
Claude3.5 explained: ``This respondent's feedback is overwhelmingly positive. They describe the project as `Accessible', `Comprehensive', `Fun', `Useful', and `Valuable'. Their explanations emphasize how enjoyable and beneficial the project was, even calling it one of their favorite projects. The high score reflects the consistently positive feedback with no negative aspects mentioned.''

An example where GPT4o (0.09), GPT4o-mini (.15) and Claude3.5 (0.20) all expressed high confidence for a negative sentiment datum was for a PRTG with sentiment of 0.20. One GPT4o explanation was: 
``The overall sentiment is highly negative due to the recurring themes of difficulty,  frustration,  and substantial time investment,  supported by detailed explanations of the challenges faced.'' (run3). 
Claude3.5 explained ``This respondent's feedback is predominantly negative. They describe the project as `Complex', `Frustrating', `Hard to use', `Stressful', and `Time-consuming'. Their explanations highlight difficulties with the IDE, lack of code comments, and unclear expectations. The low score reflects the consistently negative feedback with no positive aspects mentioned.''

Consistent medium confidence was exemplified on a PRTG with sentiment 0.50, where GPT4o averaged 0.54 and Claude3.5 rated 0.50. A GPT4o explanation was: 
``The words chosen imply a mixture of both positive and negative sentiments. Terms like `fun' imply positivity and enjoyment,  while words like `confusing' and `time-consuming' introduce negative experiences. Given the explanations provided,  the sentiment leans slightly more positive when considering the challenge and comprehensive understanding gained over time. 
Claude3.5 explained: ``This respondent's feedback is mixed. While they found the project `Fun' and `Comprehensive', they also described it as `Complex', `Confusing', and `Time-consuming'. They appreciated the challenge but noted the difficulty in understanding the program initially. The moderate score reflects this balance of positive and negative aspects, with the fun factor slightly outweighing the challenges.''

On average, GPT4o expressed high confidence in its sentiment score for 200 of the 250 word/explanation pairs; medium for 47 pairs, and low confidence for 3 pairs for the ZORQ data. For 145 pairs, all three runs expressed high confidence in their scores, and matched medium confidence for 10 pairs. 
Only run\textsubscript{2} had any low confidence ratings, and only for two sets of respondent data where the user had not provided an explanation for the words selected. In those cases, the algorithm explained ``Positive but no context provided to adjust.''

\section{Discussion and Future Work} \label{discussion}

\subsection{Discussion of Findings}
The results presented in Tables \ref{tab:sentimentresultsStats} and \ref{tab:CARMAsentimentresultsStats} demonstrate that the LLMs performed well to quantify implicit user sentiment of PDT data using both respondent-level and Avg5 scoring. This research is an initial step toward the creation of a tool for a broad user-base designed to provide rich quantitative sentiment analysis of implicit product desirability, especially in situations where no user rating system exists. 

Table \ref{tab:sentimentresultsBins} and Figure \ref{PRTGResults} provide the sentiment distributions from strongly negative ($x<0.2$) to strongly positive ($x>=0.80$), for each tool on the ZORQ data. These values show the strength of the LLMs, as compared with the other tools. Numerical sentiment as shown in Table \ref{tab:sentimentresultsStats} for ZORQ and Table \ref{tab:CARMAsentimentresultsStats} for CARMA provides a much more refined approach than categorical sentiment, that can be statistically verified for accuracy, and supported with strong MSD, MAD, PC and Avg values.

The study found that the PDT data, especially at the respondent level, provided enough data for LLMs to perform well, without domain-specific training. As a note on the applicability of the PDT as a survey tool for use in studying the effectiveness of gamification applications, it was quick and easy to construct and distribute and has a strong foundation in software product evaluation \citep{Hastings}. While the data sets used were small, there was high inter-annotator agreement, and the methods were replicated across two datasets.

Tables \ref{tab:model_comparison_4a} and \ref{tab:model_comparison_4b} provide an overview of categorical sentiment analysis of the ZORQ data. While the three class approach provided strong accuracy and F1-scores, it provides limited knowledge related to the respondent's actual sentiment. 

When looking at how much data GPT4o and GPT4o-mini need for quality results, Figures \ref{averageZORQ} and \ref{averageCARMA} along with Tables \ref{tab:zorq_stats} and \ref{tab:carma_stats} demonstrate that GPT4o-mini and GPT4o can produce statistically significant sentiment with limited PDT data, but the small amount of text in the word/explana\-tion pairs may have made it challenging for the tools to match sentiment in every run, as noted by \citet{Hartmann}. However, 
GPT4o and GPT4o-mini both had good MAE, MSE, PC values on all tests, and all runs were statistically significant with respondent scoring (Tests A and B) on the ZORQ data. The results indicate that these LLMs have been able to overcome the challenge of conducting sentiment analysis on short snippets of text. 
Further work is needed to investigate the impact of data length on LLM performance. Additionally, reproducibility is limited due to API instability and proprietary model changes (e.g., Claude vs GPT).

During initial prompt development, GPT4 through the web interface struggled to produce consistent behavior. Requesting a sentiment score appeared to help with GPT4’s behavior. Collectively, a confidence score and an explanation of scoring provided by the LLMs help in understanding user sentiment. Future work could investigate the quantitative effect on sentiment accuracy of prompting for a confidence score. A final observation is that GPT4 through the web interface was cumbersome, and even through the API it had challenges. In general, for any LLM, one-off processing of smaller PDT data through the web interface might suffice. To scale, the API is needed.

\subsection{Practical Implications}

\subsubsection{An Unbiased and Easy to Construct and Distribute Survey Method for Product Evaluation}
The process of using the PDT tool as the survey method along with LLM sentiment analysis provides for product evaluation where higher-level impressions of that product are desired. A survey using this approach is quick and easy to construct and
distribute, and is thus especially relevant to practitioners
who are short on the staff or time needed to design and distribute a survey \citep{Hastings}.

Our results confirm  the promising novel method of quantitatively measuring implicit user desirability with statistically significant results introduced in \citep{weitl2024analyzing}. The combined use of the PDT and LLM sentiment analysis also limits investigator and selection biases \citep{Volo,packer}, while providing a controlled and structured approach that is also open-ended and flexible. The PDT survey method demonstrates a simple anonymous online survey approach which can elicit candid non-biased feedback from participants about a product and should be generally applicable to a wide range of products.

\subsubsection{Refined and Accurate Sentiment Analysis}
Using prompts such as shown in Table \ref{tab:llm_prompts}, LLMs can quickly be used to provide high quality numerical and categorical sentiment analysis. When comparing the previous sentiment analysis conducted on the data sets, the LLMs allow for a much more refined use of the PDT data for quantifying and classifying sentiment, without the need for the data to be analyzed and hand classified by human experts. 

This method integrates implicit sentiment from qualitative un-rated PDT product reviews and transforms them into quantitative results. Additionally, the approach is a zero-shot method, meaning that it processes these reviews without any specific training. The LLM models offer deeper sentiment understanding than traditional ratings (e.g., Likert scales) by utilizing richer qualitative input and accurately generate nuanced sentiment scores directly from the qualitative input.

For example, when the CARMA data was first reported in 2010 \citep{Hastings}, the results were  provided through a word cloud, and by counting how often each PDT term was selected. To provide a quantitative or classification summary of the user experience, the authors had to classify by hand the individual responses of the participants. The authors classified the words selected into three categories based on the terms used and the survey context: positive (e.g., straightforward), negative
(e.g., complex), and neutral (e.g., simplistic). Using this method, CARMA’s interface was described as positive in 68\% of the words selected, negative in
25\%, with the remainder as neutral. The authors also hand classified the explanations, and noted that through the respondent explanations, CARMA’s interface was described positively in 73\% of the responses, negative in 25\%, with 2\% neutral. To use the CARMA data for product development and improvement, the original authors had to hand review the data, and noted that \textit{"The comments in some cases will prove helpful in improving CARMA’s look. While it would be time prohibitive to wade through comments in the event of a larger number of responses, such analysis was easily manageable in this case."}

When the ZORQ PDT data was first reported in 2023 \citep{Weitl}, LLM usage was just beginning. At the time,  the process of lexical sentiment analysis, which relied on tokenization, stemming, lemmatization, and the removal of stopwords in the textual information, was used. As done with the CARMA data, the authors hand counted positive, neutral and negative terms and explanations, and found that 79\% of the terms selected were positive terms, 11\% were neutral terms, and 11\% were negative terms. 

To evaluate the PRTGs as a whole, unigram lexical sentiment analysis was used on the ZORQ data in \citep{Weitl}, and they reported 58\% positive PRTGs, 20\% neutral, and 22\% negative PRTGs, using that method. Unigram lexical sentiment analysis struggles with mixed sentiment. Their results showed much fewer positive results on the same data reported above using three categories, where we found the ZORQ data had 76\% positive PRTGs, 18\% neutral PRTGs, and 6\% negative PRTGs.  The utilization of a unigram lexical approach did not take into account the syntactic composition of each response. For example, in one response, the word "fun" was chosen, with the explanation, "It was fun to problem solve and find why my ship was not doing what it was suppose to.” While the response is innately
positive in nature, the resulting unigram sentiment score rated
the response as negative, scoring it at -0.083. The negative words problem find "why" and "not" outscored the positive terms "fun" and "solve"\citep{Weitl}. Our approach of using LLM sentiment analysis does a much better job of handling mixed sentiment than the analysis methods used in \citep{Weitl}.

\subsubsection{Cost Effective Solutions}

This study assesses whether GPT-4o-mini can deliver comparable sentiment analysis accuracy while reducing resource consumption. Table \ref{tab:gen_costs} provided an analysis of the costs to run both GPT4o and GPT4o-mini, and showed that GPT4o-mini was much more cost effective, especially when working with large datasets. GPT4o-mini represents a 94\% savings over GPT4o. Although the cost to produce the GPT4o for a single run of PDT data might be acceptably low, the estimated time and cost to produce such a test on numerous occasions with new datasets, with expected results not any better than the GPT4o-mini results, may or may not be worth the cost depending on the criticality of such an effort, with the GPT4o-mini a more cost-effective choice with similar results.  

Based on the results, GPT4o-mini (Test8) would serve well as a cost-effective and highly accurate algorithm as the basis for this tool. A combination of cost-effective LLMs could also be used, similar to weighted voting or other bagging-like \citep{bagging} machine learning methods to potentially improve statistical results, without adding the cost of the larger LLMs.

\subsubsection{Confidence in Results}
The confidence and explanations expressed by the LLMs
add value in understanding user sentiment. For example, when  GPT4o, GPT4, and Claude3.5 agree in high confidence on their
ratings, the authors’ confidence in the rating increases, whereas
medium or low confidence suggests the potential need for
human review. Because the LLMs expressed high confidence a
vast majority of the time (GPT4o and GPT4 84\%; Claude3.5
62\%), human review is likely only needed for a few cases.
Further exploration is needed for confirmation. Further exploration is needed for confirmation. 

Overall, these results show that GPT4o-mini gives comparable results to GPT4o, especially when provided with the complete PRTG data from each respondent. Both models showed enough accuracy in each case (ZORQ and CARMA) to indicate that they could be used instead of human sentiment analysis if provided complete data sets.

\subsubsection{Explainable AI}

While this research is not formally framed as explainable AI (xAI), the model-generated explanations and confidence scores align well with frameworks for interpretability and transparency in AI systems \citep{doshivelez2017rigorous} for quantifying sentiment in product reviews. These explanations and confidence scores  support user trust and targeted reviews \citep{Gunning_Aha_2019}, particularly in mixed or ambiguous sentiment cases. Future work could more formally evaluate the role of these explanations in promoting user trust and system transparency.

\subsubsection{Product Development and Improvement}
One of the key practical implications of accurately understanding customer sentiments expressed in online product reviews is the opportunity it provides for product development and improvement \citep{Bharadwaj}. 
The use of the PDT survey and LLM for sentiment analysis also provides easy ways to identify ideas for product development and improvement, as well as marketing ideas for their target audience. With an LLM, implicit sentiment from written reviews can be integrated into quantitative scores that give comprehensive information in a scalable manner.

The words selected by the respondents and the explanations they give provide rich data for product development and improvement. 
By analyzing the sentiment and explanations, businesses can gain valuable feedback on their products and identify areas for enhancement \citep{Bharadwaj}. 

Positive sentiments expressed can provide valuable insights into the features or aspects of the product that resonate well with customers, and potentially be used to create marketing messages. By identifying positive sentiments and user explanations, businesses can understand the strengths of their products and leverage the PDT explanations to enhance future iterations, further enhancing customer satisfaction and fostering positive relationships \citep{Bharadwaj}. By incorporating positive customer PDT explanations into their marketing materials as testimonials, businesses can build trust, credibility, and enthusiasm among potential customers.

By aligning future product iterations with user explanations, particularly the negative responses, businesses can discover areas where the product may be falling short of customer expectations. Ultimately, by leveraging sentiment analysis insights, businesses can enhance overall customer satisfaction, increase loyalty, and drive long-term success \citep{Bharadwaj}.

Furthermore, by monitoring and analyzing customer sentiments over time, businesses can track the impact of their product improvements, measure customer satisfaction levels, and identify recurring patterns or themes in customer sentiments. This iterative feedback loop allows businesses to refine their products based on real-time customer insights, ensuring that they remain competitive and responsive to changing market demands \citep{Bharadwaj}, and is future work for this research. 

\subsection{Limitations and Future Work}

This work was conducted on two small PDT data sets, which is a limitation of the findings. Future work would need to be verified on other and potentially larger datasets. While it is expected that this approach is applicable to other PDT data sets, including data related to non-software product reviews, further exploration is needed and would add verification and support. In designing a study using PDT data sets, care should be taken to determine the proper sample size statistically, or the full population used otherwise.  Future work is also needed which uses non-researchers to provide HITL gold-standard annotation. This could be done as new data sets are explored. Additionally, research should consider ethical implications prior to applying this method to sensitive domains such as education and healthcare.

Further work is needed to investigate the impact of data length on LLM performance. However, reproducibility is limited due to API instability and proprietary model changes (e.g., Claude vs GPT). 

A deeper exploration of the approaches for evaluating software desirability used in this paper as a generalized methodology is needed. One area of exploration is the algorithms' ability to match at a given context-level. 

The selected LLMs (GPT4o and Claude 3.5) represent two advanced LLMs \citep{Krugmann}, but our model is adaptable for use with other LLMs. Future work would be testing other LLMs, especially cost-effective versions such as Gemini Flash 2.0~\citep{Gemini}, as well as on ensembles of cost-effective LLMs.

The confidence and explanations expressed by the LLMs add value in understanding user sentiment. For example, when GPT4o-mini, GPT4o, GPT4, and Claude3.5 agree in high confidence on their ratings, the authors' confidence in the rating increases, whereas medium or low confidence suggests the potential need for human review. Because the LLMs expressed high confidence a vast majority of the time (GPT4o, GPT4o-mini and GPT4 84\%; Claude3.5 62\%), 
human review is likely only needed for a few cases. Further exploration is needed for confirmation. 

Overall, these results show that GPT4o-mini gives comparable results to GPT4o, especially  when  provided with the complete PRTG data from each respondent.  Both models showed enough accuracy in each case (ZORQ and CARMA) to indicate that they could be used instead of human sentiment analysis if provided complete data sets.

In general, for any LLM, one-off processing of smaller PDT data through the web interface might suffice. To scale, the API or even a full application is needed. Our model formalizes a process that takes PDT qualitative data as input, uses LLMs internally to create numeric sentiment scores along with confidence explanations for user evaluation that can then be applied to decision making.

While this research is not formally framed as explainable AI (xAI), the model-generated explanations and confidence scores align well with frameworks for interpretability and transparency in AI systems \citep{doshivelez2017rigorous}. These elements may support user trust and targeted review \citep{Gunning_Aha_2019}, particularly in mixed or ambiguous sentiment cases. Future work could more formally evaluate the role of these explanations in promoting user trust and system transparency.

We do not conduct extensive prompt engineering, so there may likely be better prompts to obtain better performance. This is left to future work.

 \section{Conclusion} \label{conclusion}
 In general, this research provides a generalizable pipeline for converting qualitative user experience data into quantitative sentiment scores, and provides a new evaluation paradigm for PDT-style instruments. This research adds to a deeper understanding of evaluating user experiences. Using the PDT tool as the survey method along with cost efficient LLMs for sentiment analysis has the potential to provide for quick product evaluation with limited investigator bias. The results are rich in terms of sentiment scores (both numerical and classified sentiment) as well as with high-level impressions of the product for further analysis. It also provides an easy way to identify ideas for product development and improvement, as well as marketing ideas for target audiences. 
 
This research explores the use of several LLMs (Claude Sonnet 3 and 3.5, GPT4, GPT4o and GPT4o-mini) along with TRBS and VADER on sets of PDT data, for providing both classified and quantitative numerical zero-shot sentiment analysis of implicit software desirability expressed by users. All LLM tools outperformed the other approaches and were statistically significant in performing as zero-shot sentiment analyzers on the PDT data. Numerical sentiment analysis fit the data better than categorical sentiment analysis and provided much finer analysis. 

GPT4o-mini achieved comparable performance to GPT\-4o, at a 94\% lower cost.  Both maintained robustness even when handling data presented in multiple forms and consistently expressed high confidence with human-readable justifications. The confidence and explanation of confidence provided by the LLMs assist in understanding the user sentiment.


\section{Declaration of generative AI and AI- assisted technologies in the manuscript preparation process}

During the preparation of the manuscript, the authors used ChatGPT in order to perform an evaluative review, and for slight wording adjustments in the abstract and conclusion, and to assist with finding grammatical/punctuation mistakes. After using this tool/service, the authors selected to slightly revise the abstract and conclusion, and corrected the grammatical/punctuation mistakes that were found. The authors take full responsibility for the content of the published article.

\balance
\bibliographystyle{cas-model2-names}

\bibliography{references}


\bio{}
Sherri Weitl-Harms is an Associate Professor of Computer Science at Creighton University. In addition to several years of industry experience, she has been in academia for more than 25 years and served as chair of the Cyber Systems department at the University of Nebraska at Kearney for 12 years, leading its efforts to obtain ABET accreditation. Her research areas include machine learning/artificial intelligence, natural language processing, CS education, gamification, health informatics, and spatial-temporal data mining. She is a highly productive researcher (producing 40+ peer-review publications and contributing to projects totaling more than \$12M in external funding), a well-respected educator (teaching more than 20 different courses and mentoring hundreds of undergraduates), and a leader both in her institution and in her professional discipline. Dr. Weitl-Harms is heavily involved in undergraduate research, serves as a Division Representative for Math/CS (MCS) on the Council for Undergraduate Research (CUR), mentoring numerous student publications and presentations,  and received the 2025 Advanced-Career Faculty Mentor Award for the MCS Division of CUR. She is actively committed to service learning, with student projects that have aided hundreds of local, regional, and national organizations. She is an editor, program committee member and reviewer for several professional conferences and journals, a Program Coordinator for the Nebraska and Southwest Iowa NCWIT Aspirations Alliance, and a Senior IEEE member. Dr. Weitl-Harms received the 1997 Governor’s Award for Teaching Excellence for Lincoln University in Missouri.  She has led and participated in numerous CS educational outreach activities including Cyber camps, Coder Dojos, Hour of Code events, and judging FBLA events. She is a member of ACM, IEEE, Computer Science Teachers Association, NCWIT Academic Alliance, the Gamma Omicron chapter of Epsilon PI Tau, Anita Borg Society, Women in Cyber Security and the National Education Association. She received her PhD from the University of Missouri.
\endbio

\bio{}
Dr. John Hastings is a Professor in The Beacom College of Computer \& Cyber Sciences. He has over 35 years of teaching and curriculum development experience in computer science, artificial intelligence, and cybersecurity related courses at both the undergraduate and graduate level. 
His research focuses on machine learning and artificial intelligence including natural language processing (LLMs), generative AI, computer vision, ecological and environmental applications of AI, AI in games, cybersecurity, and gamification in CS and cybersecurity education. He has won several awards related to his research. He has several years of industry experience as an AI/ML software engineer, team leader, and as a business owner providing AI/ML consulting services. He received his PhD from the University of Wyoming.
\endbio

\end{document}